\documentclass{article}

\usepackage{PRIMEarxiv}

\usepackage{abstract}
\usepackage[utf8]{inputenc}
\usepackage[T1]{fontenc}
\usepackage{hyperref}
\usepackage{url}
\usepackage{booktabs}
\usepackage{array}
\usepackage{amsfonts}
\usepackage{amsmath}
\usepackage{amsthm}
\usepackage{nicefrac}
\usepackage{microtype}
\usepackage{fancyhdr}
\usepackage{graphicx}
\usepackage{placeins}
\usepackage{subcaption}
\graphicspath{{media/}}

\newtheorem{proposition}{Proposition}[section]

\newcolumntype{L}[1]{>{\raggedright\arraybackslash}p{#1}}

\title{Anomaly Detection from a Tensor Train Perspective}

\author{
  Alejandro Mata Ali \\
  Instituto Tecnológico de Castilla y León, Burgos, Spain\\
  \texttt{alejandro.mata.ali@gmail.com} \\
   \And
  Aitor Moreno Fdez. de Leceta \\
  Quantum Technologies and Systems Unit,\\
  LKS Next, MONDRAGON Corporation, Goiru 7,\\
  20500 Arrasate-Mondragón, Gipuzkoa, Spain\\
  \texttt{aitormoreno@lksnext.com} \\
   \And
  Jorge López Rubio \\
  i3B Ibermatica, Parque Tecnológico de Bizkaia \\
  Ibaizabal Bidea, Edif. 501-A \\
  48160 Derio, Spain\\
  \texttt{jlopezru@ayesa.com} \\
}

\begin{document}
\maketitle

\begin{abstract}
We present a series of algorithms  in tensor networks for anomaly detection in datasets, by using data compression in a Tensor Train representation. These algorithms consist of preserving the structure of normal data in compression and deleting the structure of anomalous data. The algorithms can be applied to any tensor network representation. We test the effectiveness of the methods with digits and Olivetti faces datasets and a cybersecurity dataset to determine cyber-attacks.
\end{abstract}

\section{Introduction}
Anomaly detection is a task with a wide range of applications, ranging from industrial monitoring \cite{Industrial_Anomaly} and medical diagnostics \cite{Medical_anomaly} to fraud detection \cite{Fraud_Anomaly} and cybersecurity \cite{Ciber_anomaly}. The primary goal of anomaly detection is to identify data or structures that deviate significantly from normal behavior, often serving as an indicator of a cyberattack, fraud, system malfunction, or other unusual events. Traditionally this task has been performed with statistical methods \cite{Statistical_Anomaly}, machine learning \cite{ML_Anomaly} or deep learning \cite{Deep_Anomaly}. Although these methods achieve good results, they have the limitation of not being able to adequately deal with high dimensionality data, which is usually addressed with dimensional reduction techniques such as Principal Component Analysis (PCA) \cite{PCA}.

Recently, quantum computing has gained great popularity for its promising capabilities, and with it, quantum-inspired technologies. One of the best known technologies is tensor networks (TN) \cite{TensorNetwork}, which consist of graphical representations of multidimensional linear algebra operations. This set of techniques allows to deal with quantum system simulations and general problems of high dimensionality in an efficient way, either through reformulations of the problem or through approximate representations such as Matrix Product State or Tensor Trains (MPS/TT) \cite{MPS1,Oseledets2011} or Projected Entangled Pair States (PEPS) \cite{MPS2}.

Tensor networks have already been applied to the context of anomaly detection \cite{TN_Anomaly}, in a supervised manner, imitating the structure of autoencoders, taking advantage of the compression power of Matrix Product Operator (MPO) representations for certain transformations of the input data.

The approximate representations allow to keep the most relevant structures of the tensors they represent, eliminating the less relevant ones and reducing the space required for their storage. The different representations allow to keep different types of structures efficiently and to recover the original tensor with different contraction schemes. In a dataset, normal data usually present one or several structures that relate them to each other, either because they are very similar or because the structure is repeated very frequently, while anomalous data have a different structure from each other, or are infrequent with respect to the normal dataset.

We present an anomaly detection method based on compressing the dataset into an approximate representation, in a controlled manner, such that normal data are less perturbed by compression than anomalous data. As a consequence, the normal data will be less displaced from their initial position than the anomalous data, allowing them to be distinguished by performing the scalar product of each original data with its compressed version. We have 8 different algorithms available based on 2 conceptually different compression methods, one that compresses the whole dataset and one that compresses each individual data point. For simplicity, we present the algorithms with a TT representation, but it is simply generalizable to any other representation.

\section{Compression Tensor Network Anomaly Detector}
Given a set of data expressed as a matrix $A$, whose rows store each data point and its columns each data feature, in which there is a subset of normal data and a subset of anomalous data, our objective is to distinguish these two subsets. Normal data follows a given distribution (or structure), which may or may not be given previously, and is more frequent than anomalous data. Anomalous data does not fit this distribution and is less frequent than normal data, usually having a more diffuse and less well defined distribution (or structure) than normal data.

In order to understand the algorithms we are going to present, we first briefly discuss data compression using tensor network representations.

\begin{figure}
    \centering
    \includegraphics[width=0.5\linewidth]{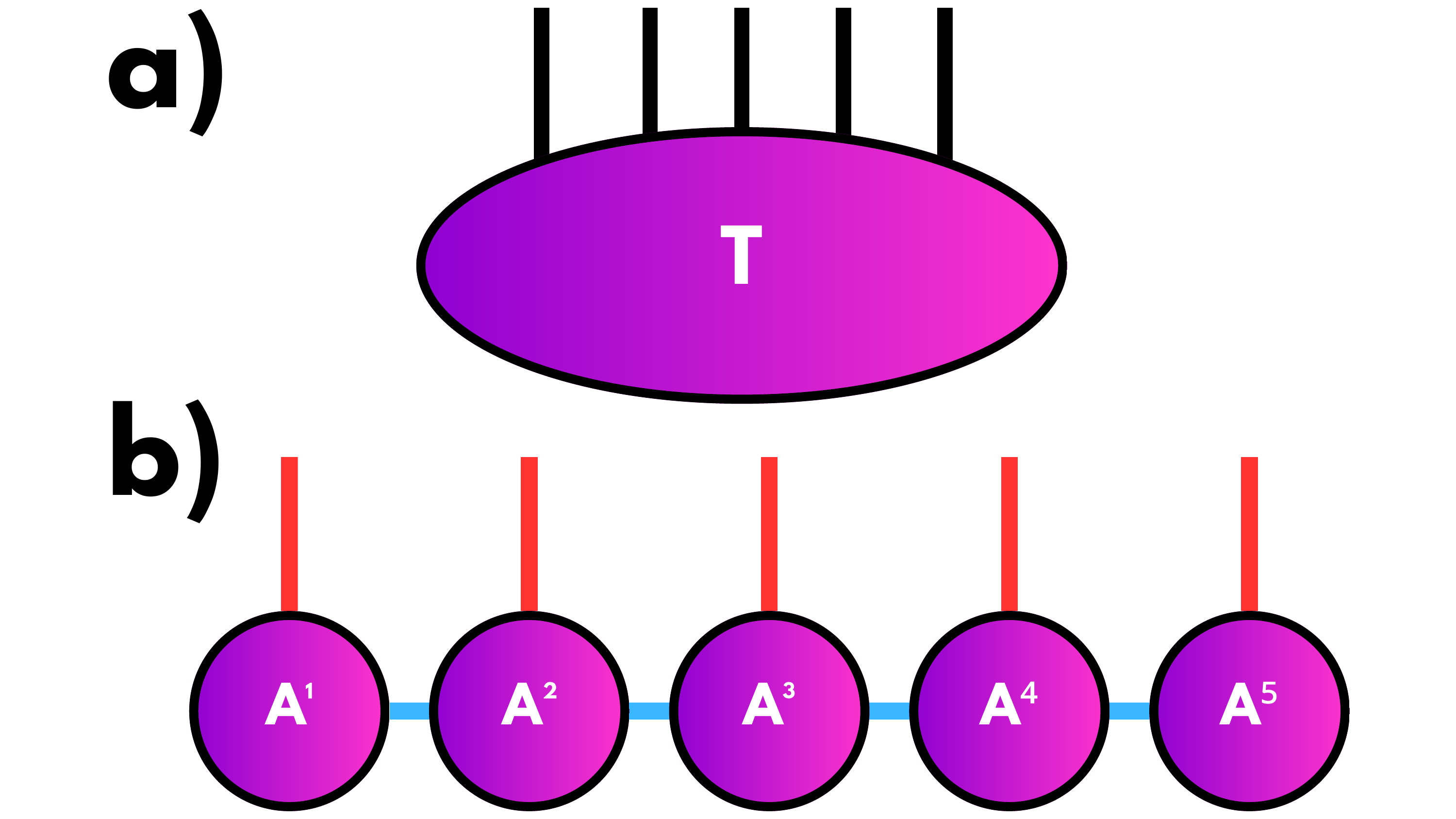}
    \caption{a) Tensor $T$ of order 5 in tensor networks notation, b) TT representation of a tensor of order 5 in tensor networks notation}
    \label{fig:Tensor_TT}
\end{figure}

If we have a tensor $T$ of order $n$, with elements $T_{i_1,i_2,...i_n}$, it is represented as a node with $n$ outgoing lines, each one representing an index. We can see an example in Fig. \ref{fig:Tensor_TT} a.

This tensor can change its order by splitting or grouping its indexes, increasing or reducing the number of indexes respectively.  In grouping, we reduce the number of indexes by a bijective mapping from the initial index set $i_1, i_2, \dots, i_n$ to a final index set $j_1, j_2, \dots, j_m$. If the initial index set had dimensions $\vec{d}$, the final set has a set of dimensions $\vec{d}'$ such that
\begin{equation}
    \prod_{k=1}^n d_k = \prod_{k=1}^m d'_k.
\end{equation}

A tensor of order $n$ can be transformed to an TT representation of the same tensor consisting of 2 tensors of order 2 and $n-2$ tensors of order 3, in the form of a linear chain contracted through the indexes that we call bond indexes, with a free index for each index of the original tensor, which we call physical indexes. The equation we have is
\begin{equation}
    T_{i_1,i_2,\dots,i_n} = \sum_{j_1,j_2,\dots,j_{n-1}} A^{1}_{i_1,j_1} A^{2}_{j_1,i_2,j_2}\dots  A^{n}_{j_{n-1},i_n} 
\end{equation}

We can see in Fig. \ref{fig:Tensor_TT} b an TT representation with the red indexes being the physical ones and the blue indexes being the bond ones. There are more types of representations with different structures, such as 2-D PEPS or hierarchical trees.
The representations can be exact, so that when contracted they return exactly the tensor represented, or approximate, returning a tensor as similar as desired by controlling the degree of approximation. In the approximation process the tensor is compressed, reducing the dimension of the bond indexes, preserving a smaller number of internal structures of the tensor. This can be done both for a representation already created and in the process of creating it. 

\begin{figure}
    \centering
    \includegraphics[width=0.5\linewidth]{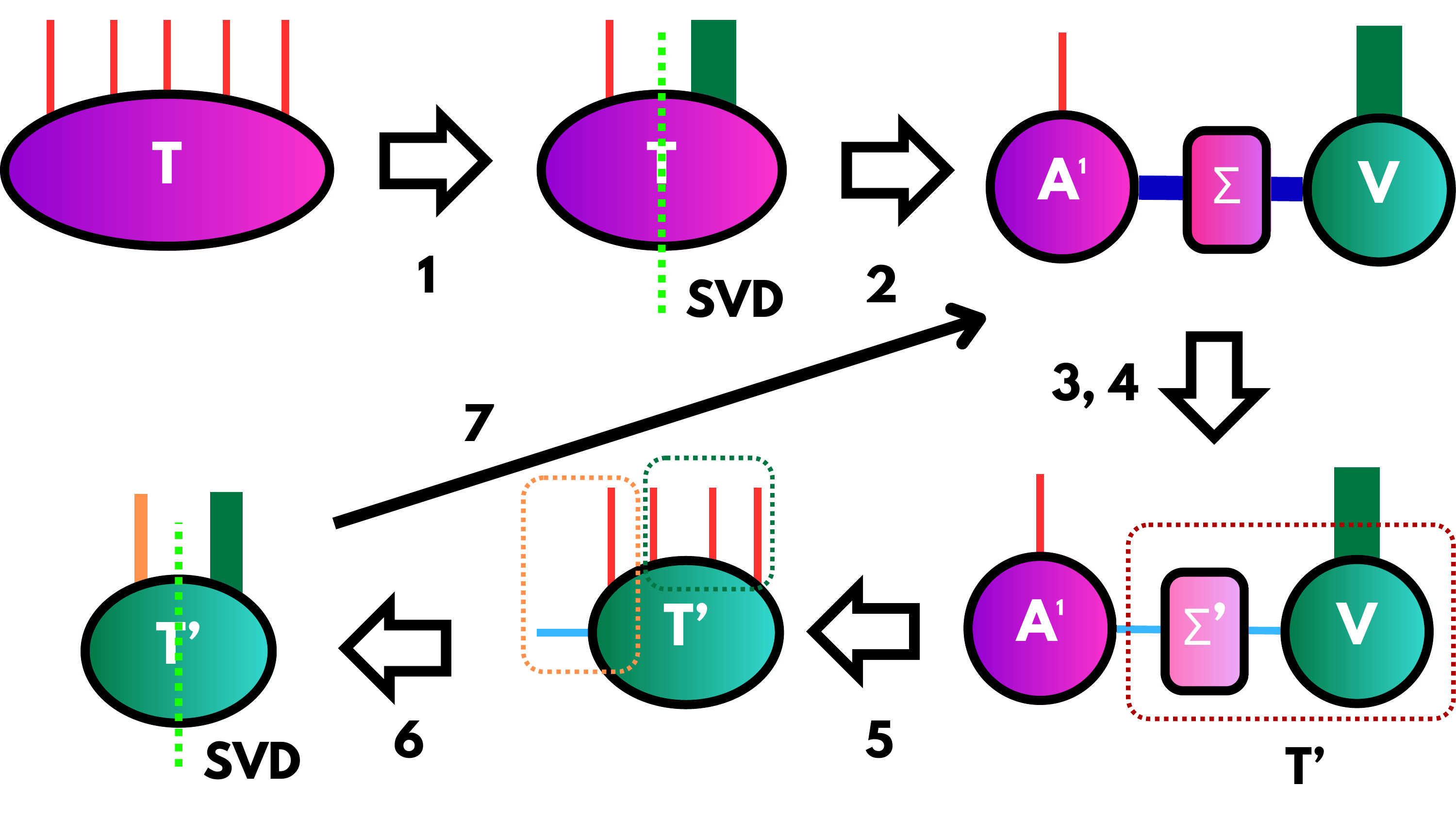}
    \caption{Process of creation of the TT representation for a tensor of order 5, following the described steps.}
    \label{fig: TT SVD}
\end{figure}

The creation of a TT representation can be accomplished in several ways, but a simple method that is often the best approach is the iterative use of singular value decompositions (SVD), often referred to as TT-SVD \cite{Oseledets2011}. This method offers a systematic approach to how to obtain an approximate representation in a controlled manner by means of a compression parameter $\tau$, which is what we need for our algorithms. The process is as follows in Fig. \ref{fig: TT SVD}.

\begin{enumerate}
    \item We perform a grouping of all the indexes from the second to the last one, having a tensor of order 2.
    \item We perform a SVD to this tensor of order 2, obtaining 3 matrices $U$, $\Sigma$, $V$, so that $U$ will be our $A^1$, and $\Sigma$ is a diagonal matrix with the singular values $\sigma_k$ of the decomposition.
    \item Let $\sigma_1=\sigma_{\max}$ be the largest singular value. We keep only the singular values that satisfy $\sigma_k \geq \tau\sigma_1$, so we discard the least relevant information.
    \item We keep the rows and columns of $\Sigma$ associated with the singular values we have kept, just as for $V$ and $U$, being our new $U$ the $A^1$ tensor.
    \item We multiply $\Sigma$ with $V$ to get the tensor $T'$ of the next iteration.
    \item We split all of the indexes of $T'$ except the one that bonds it with $A^1$ and we do grouping of the second physical index (the next one from which we need a tensor) and the index obtained in the SVD, and also grouping of the other indexes.
    \item We perform the SVD for this tensor and truncate again as in steps 3, 4 and 5.
    \item We repeat the process until we obtain the whole TT representation.
\end{enumerate}

Once this process is done, we have the TT representation of the tensor with a compression parameter $\tau$. Let $\sigma_1=\sigma_{\max}$. For $\tau\in[0,1]$, the retained rank at each truncation step is
\begin{equation}
    r_\tau=\#\left\{k:\sigma_k\geq\tau\sigma_1\text{ and }\sigma_k>0\right\},
\end{equation}
or, in numerical terms,
\begin{equation}
    r_\tau=\#\left\{k:\sigma_k\geq\tau\sigma_1\text{ and }\sigma_k>\varepsilon_{\mathrm{mach}}\sigma_1\right\}.
\end{equation}
Thus, $\tau=0$ retains the full numerical rank, that is, all positive singular values, corresponding to minimal compression by thresholding. At the other extreme, $\tau=1$ retains only the singular values equal to $\sigma_1$, so at least one singular value is kept whenever the unfolding is nonzero. As $\tau$ increases, the number of retained singular values cannot increase and typically decreases; therefore, larger $\tau$ means stronger compression. The parameter $\tau$ can also be a list of compression factors, so that each truncated SVD has a different factor to compress. We take it as a constant in the remainder of the paper.

Having this formulation explained, we can proceed to the central algorithms of the paper.

\subsection{Global Compression Algorithm}
For the global compression method, we use as tensor the global dataset expressed as a $A_{ij}$ element matrix of $N\times M$ dimensions, having $N$ data points with $M$ features. We perform splitting to the matrix on the index of the columns, so that we obtain a tensor of order $n$ on which to perform the compression in TT representation. In case we do not have a suitable dimension, we can add 0 at the end of each data so that we have a $N\times M'$ matrix $A'$. We add 0 to avoid altering the norm or the structure of the data.

We apply the compression in TT representation with the desired compression factor $\tau$, obtaining a tensor network that represents the initial data in a compressed form. In this way, the most normal data, sharing a common set of structures, can still be represented accurately even with an approximate representation. On the other hand, anomalous data, having different structures among them, will be seen in the representation as details, and when compressing the representation, it will no longer be represented efficiently and will be displaced from their initial position. In addition, if they appear less frequently, or have a weaker structure than normal data (for example, if the anomalous data tend to have a lower norm), the structure of the anomalous data will disappear much more easily, shifting even more. 

To determine how much the original data has been displaced with respect to the compressed data in this representation, we have many options on how to calculate the decision function. In the self-comparative case we choose the self-retention score
\begin{equation}\label{eq: decision global auto}
    s_{\mathrm{self}}(y,\hat y)=\frac{\langle y,\hat y\rangle}{\lVert y\rVert_2^2},
\end{equation}
where $y$ is the original vectorized datum and $\hat y$ is its compressed reconstruction. This is not a pure cosine similarity. If $\theta$ denotes the angle between $y$ and $\hat y$, then
\begin{equation}\label{eq:self-retention-factorization}
    s_{\mathrm{self}}(y,\hat y)=\frac{\lVert \hat y\rVert_2}{\lVert y\rVert_2}\cos(\theta).
\end{equation}
Hence the self-comparative score combines directional alignment and magnitude retention. In particular, if $\hat y=\alpha y$, then $s_{\mathrm{self}}(y,\hat y)=\alpha$, so even when the direction is unchanged, a change in magnitude changes the score. This is desirable in the self-comparative setting because magnitude loss or amplification after compression may itself indicate anomalous behaviour.

A complementary quantity is the relative reconstruction error
\begin{equation}\label{eq:self-relative-error}
    e_{\mathrm{self}}(y,\hat y)=\frac{\lVert y-\hat y\rVert_2^2}{\lVert y\rVert_2^2}
    =1+\frac{\lVert \hat y\rVert_2^2}{\lVert y\rVert_2^2}-2\frac{\langle y,\hat y\rangle}{\lVert y\rVert_2^2}.
\end{equation}
We do not replace the reported score by this error; we mention it only to clarify the interpretation of \eqref{eq: decision global auto}.

The TT-SVD compression map is built by sequential truncated SVD steps and should therefore not be identified, in full generality, with a single global linear orthogonal projector. Nevertheless, the intuition behind \eqref{eq: decision global auto} can be explained through the following idealized model.

\begin{proposition}[Idealized projection model]
Let $P$ be an orthogonal projector. If
\[
    \lVert (I-P)y\rVert_2 \leq \varepsilon \lVert y\rVert_2,
\]
then
\[
    \frac{\langle y,Py\rangle}{\lVert y\rVert_2^2}
    \geq 1-\varepsilon^2.
\]
Conversely, if
\[
    \lVert (I-P)z\rVert_2 \geq \delta \lVert z\rVert_2,
\]
then
\[
    \frac{\langle z,Pz\rangle}{\lVert z\rVert_2^2}
    \leq 1-\delta^2.
\]
\end{proposition}

\begin{proof}
Since $P$ is an orthogonal projector, $P=P^\top=P^2$ and
\[
    \langle y,Py\rangle=\lVert Py\rVert_2^2=\lVert y\rVert_2^2-\lVert (I-P)y\rVert_2^2.
\]
Therefore
\[
    \frac{\langle y,Py\rangle}{\lVert y\rVert_2^2}
    =1-\frac{\lVert (I-P)y\rVert_2^2}{\lVert y\rVert_2^2},
\]
from which both inequalities follow immediately.
\end{proof}

This proposition should be read as an idealized explanation of the score, not as a universal guarantee for the full TT-SVD compression map. In this picture, normal samples are well preserved because most of their energy lies in a normal subspace, whereas anomalous samples lose more energy under compression and therefore attain lower self-retention scores.

This scalar product can be calculated either by contracting the data tensor with the representation of this tensor and keeping only the diagonal of the resulting matrix, or by contracting the representation and performing the scalar product directly for each data point.

This first option is called Auto Comparative Global Compression Tensor Network Anomaly Detector (ACGCTNAD), and its main function is that, given a set of unlabeled data, with no more information than the dataset itself, to determine for each data point in the dataset whether it is normal or anomalous with respect to the dataset itself. This would fall under the denomination of unsupervised method.

However, we can include extra information about what is normal by adding to the dataset to be compressed a set of data, which we can call training data, that we previously know to be normal data. In this way, since we already have a dataset with a certain predominant structure, when compressing the total dataset, this will be maintained over any other possible structure provided by the dataset to be tested. Thus, the decision values are calculated exactly as before, but limited to the data we want to test. This would be a supervised method.

Another alternative is to use data that is mostly normal, or we believe to be normal, as training data, since the method itself processes such data and finds the underlying structure of it. Therefore, the method is quite flexible regarding the training data, and this version would be a semi-supervised method.

The second option we can perform based on this global compression method is to compare each original data point with all the other compressed data, so that we can identify which data is the least similar to all the others and find relationships between the data. To do this, to determine if a data point $y$ is normal or anomalous we use
\begin{equation}\label{eq: decision global group}
    s_{\mathrm{group}}(y)=\sum_{i=1}^N\frac{\langle y,\hat x_i\rangle}{\lVert y\rVert_2\lVert \hat x_i\rVert_2},
\end{equation}
where $\hat x_i$ is the compressed version of the $i$-th data point of the set. Here the cosine normalization is intentional: in the group-comparative setting we compare different data points, which may have naturally different baseline magnitudes, and therefore we suppress those scale differences to focus on geometric alignment with the compressed normal structure. The difference between \eqref{eq: decision global auto} and \eqref{eq: decision global group} is thus a methodological design choice rather than an inconsistency.

To do this, it is necessary to multiply the original data matrix with the TT representation, and in the resulting matrix, normalize each projection, and then add the normalized projections for each original data point. It is important not to try to do this by normalizing the data before feeding it to the compressor and then using averaging, as this will alter the underlying structures of the mechanism.

We call this method Group Comparative Global Compression Tensor Network Anomaly Detector (GCGCTNAD). In this process we again have the options unsupervised, if we do not add data, supervised, if we add normal data, and semi-supervised, if we add mostly normal data.

\subsection{Local Compression Algorithm}
In the global case we need to process a certain amount of data each time we want to test a dataset, which for certain applications may not be ideal. Therefore, we propose another method based on the same idea, but for which we only need a normal representative from the normal dataset. This is less effective for the case where we have several sets of normal data, but this difficulty can be solved if we have one data point from each of these sets and make one of these solvers for each of these training data.

The idea is, given some training data $x$, to split its column index, as in the global method, find an approximated TT representation of that data, and then make a TT representation of each data point to be tested, so that the first $n-1$ nodes are equal to those of the training data representation. By doing this, we force the data to be tested to adapt to the structure of the training data, making the remaining node contain all the possible information of the tensor. In order to do this properly, we make the training representation left orthogonal, so that each of these tensors is left isometric.

\begin{figure}
    \centering
    \includegraphics[width=0.5\linewidth]{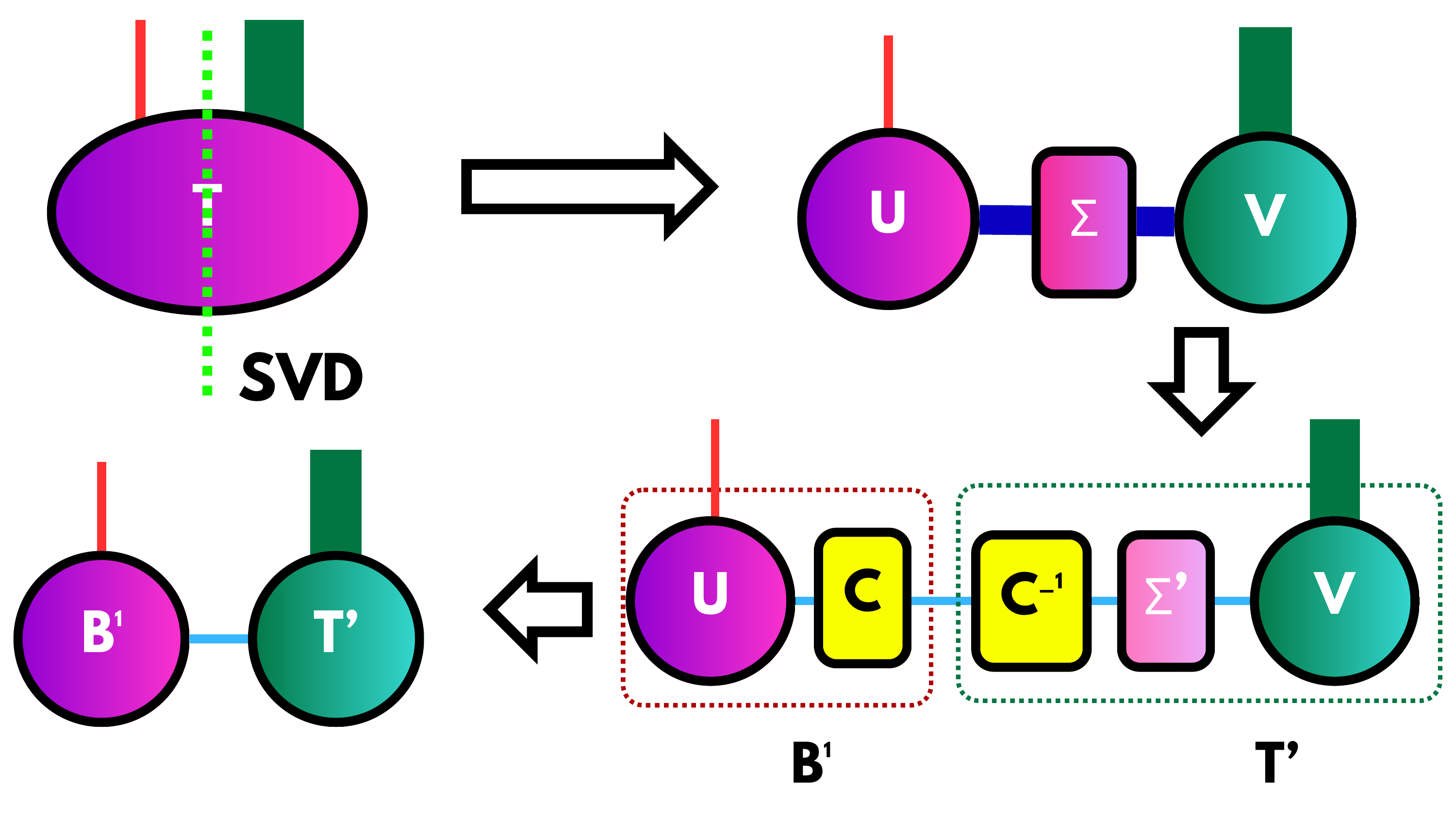}
    \caption{Example of the TT representation generation process for the data to be tested.}
    \label{fig:TT_Local}
\end{figure}

The following procedure used in the experiments should be interpreted as a local TT-core alignment heuristic. First, we obtain the left-orthogonal TT representation of the training data and keep the first $n-1$ nodes $B^i$, where $B^i$ is the node associated with the $i$-th physical index. After that, for each data point to be tested, we obtain its TT representation while forcing the first $n-1$ nodes to match the training cores. If $U$ denotes the left singular-vector factor obtained at a given truncated SVD step, one may write heuristically
\begin{equation}
    B^i=UC,\qquad C=U^{\dagger}B^i.
\end{equation}
This expression describes an alignment step between the truncated basis generated by the test datum and the corresponding normal TT core. However, in general $U$ is rectangular with orthonormal columns and $B^i$ is only left-isometric, so $C$ is not guaranteed to be square, unitary, or invertible. If the truncated bases have the same rank and span the same column space, then the corresponding change-of-basis matrix $C$ is orthogonal or unitary and one may write $C^{-1}=C^\dagger$. Outside this idealized case, this equality is not guaranteed.

Therefore, the algebraic manipulation involving $C=U^{\dagger}B^i$ should be read as a heuristic alignment device rather than as a general orthogonal projection identity. The local method used in the experiments follows this heuristic in order to compare the local TT structure induced by the test datum with the TT cores obtained from normal data. We do not reinterpret the reported numerical results as if they had been generated by a different local solver.

Once we have the TT representations of the data to be tested, we obtain the appropriate decision value. For the self-comparative local method, we compare each data point with its compressed version by means of \eqref{eq: decision global auto}, which preserves information about both directional alignment and magnitude change while keeping in memory only the $n-1$ training nodes and the datum to be tested. We call this method Auto Comparative Local Compression Tensor Network Anomaly Detector (ACLCTNAD). In this case, it is an exclusively supervised method, since we need to provide normal data representative of the dataset.

The second option is to compare the new data with the compression of all the data of a set we have, which can be all the same data we introduce or can include extra data points that we know to be normal, by \eqref{eq: decision global group}. We call this method Group Comparative Local Compression Tensor Network Anomaly Detector (GCLCTNAD). In this case the cosine normalization is again deliberate because different data points may have different natural magnitudes, and the comparison is intended to emphasize angular similarity to the compressed normal set.

\subsubsection{A projection-based local formulation}
The heuristic above suggests a mathematically principled local variant based on orthogonal projection. Let
\[
    Y\in\mathbb{R}^{d_1\times\cdots\times d_n}
\]
be a test tensor, and let
\[
    B^{(1)},\dots,B^{(n-1)}
\]
be a left-orthogonal chain of TT cores obtained from normal data. By contracting the first $n-1$ normal TT cores, we obtain an interface matrix
\[
    Q\in\mathbb{R}^{D\times r_{n-1}},\qquad D=d_1\cdots d_{n-1},
\]
with orthonormal columns, so that $Q^\top Q=I$. Let $Y_{<n>}\in\mathbb{R}^{D\times d_n}$ be the unfolding of $Y$ that groups the first $n-1$ modes as rows. The best approximation sharing the normal TT interface is then obtained from the least-squares problem
\begin{equation}
    \min_H\lVert Y_{<n>}-QH\rVert_F^2,
\end{equation}
where $H\in\mathbb{R}^{r_{n-1}\times d_n}$.

\begin{proposition}
Let $Q\in\mathbb{R}^{D\times r}$ satisfy $Q^\top Q=I_r$. For any
$Y\in\mathbb{R}^{D\times m}$, the minimizer of
\[
    \min_{H\in\mathbb{R}^{r\times m}}
    \lVert Y-QH\rVert_F^2
\]
is
\[
    H^\star = Q^\top Y.
\]
Moreover, the minimum value is
\[
    \lVert (I-QQ^\top)Y\rVert_F^2.
\]
\end{proposition}

\begin{proof}
Since $Q^\top Q=I_r$, we may decompose
\[
    Y-QH
    =
    Q(Q^\top Y-H) + (I-QQ^\top)Y.
\]
The two terms on the right-hand side are orthogonal with respect to the
Frobenius inner product. Hence
\[
    \lVert Y-QH\rVert_F^2
    =
    \lVert Q^\top Y-H\rVert_F^2
    +
    \lVert (I-QQ^\top)Y\rVert_F^2.
\]
The second term does not depend on $H$, and the first term is minimized
uniquely by $H=Q^\top Y$.
\end{proof}

Therefore the projected reconstruction is
\begin{equation}
    \hat Y_{<n>} = QQ^\top Y_{<n>},
\end{equation}
and a natural local normality score is
\begin{equation}
    s_{\mathrm{loc}}(Y)=\frac{\lVert Q^\top Y_{<n>}\rVert_F^2}{\lVert Y_{<n>}\rVert_F^2}.
\end{equation}
The corresponding anomaly score is
\begin{equation}
    a_{\mathrm{loc}}(Y)=1-s_{\mathrm{loc}}(Y)=\frac{\lVert (I-QQ^\top)Y_{<n>}\rVert_F^2}{\lVert Y_{<n>}\rVert_F^2}.
\end{equation}
Thus, $s_{\mathrm{loc}}(Y)$ close to $1$ means that the test tensor is well explained by the TT interface learned from normal data, whereas $a_{\mathrm{loc}}(Y)$ close to $1$ means that a large fraction of the tensor lies outside that interface. This formulation does not require $C$ to be square, invertible, or unitary, and it provides a direct least-squares optimality statement.

This projection-based formulation provides a mathematically consistent local alternative to the ACLCTNAD alignment heuristic used in the experiments. The numerical results reported in this work correspond to the original heuristic version; evaluating the projection-based variant is left for future work.

\section{Results}
We test these algorithms using the digits dataset \cite{Digits}, to distinguish each number from the others, Olivetti faces dataset \cite{Olivetti}, to distinguish types of faces, and also using real cybersecurity data, where we distinguish normal requests and cyberattacks. We study both ACGCTNAD and ACLCTNAD, because group comparison methods have proven to be excessively slow and with unsatisfactory results.

The first test we will perform will be on the digits dataset, for which we will choose one digit as normal class, considering the rest as anomalous. For this case we are going to use an unsupervised method, without extra normal data. We will take sets of about 150 random normal data from the dataset and the same number of outliers, in order to have a sample in which the normal data has a predominant structure. For each case in which we take one type of digit as normal, we will check the Area Under the ROC (AUROC or AUC) obtained for a series of values of the compression factor $\tau$, to determine the behavior of the anomaly detection capability of the system as a function of its fundamental parameter. The dimensions of the images are $8\times 8$, so for the splitting of the columns we choose dimensions $\{2,2,2,2,2,2\}$ for all cases, since it is the one with the maximum possible number of indices and does not require the addition of zeros. Before entering the data into the solver, we are going to apply a Standard Scaler to the data.

\begin{figure}
    \centering
    \begin{subfigure}{0.180\linewidth}
    \includegraphics[width=\linewidth]{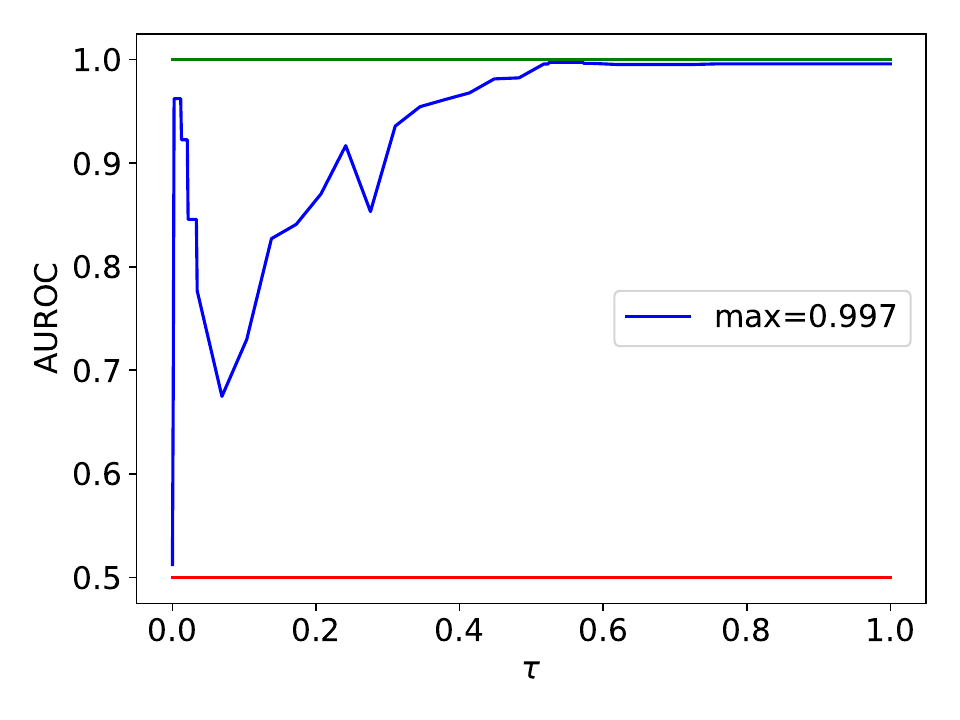}
    \end{subfigure}
    \hfill
    \begin{subfigure}{0.180\linewidth}
    \includegraphics[width=\linewidth]{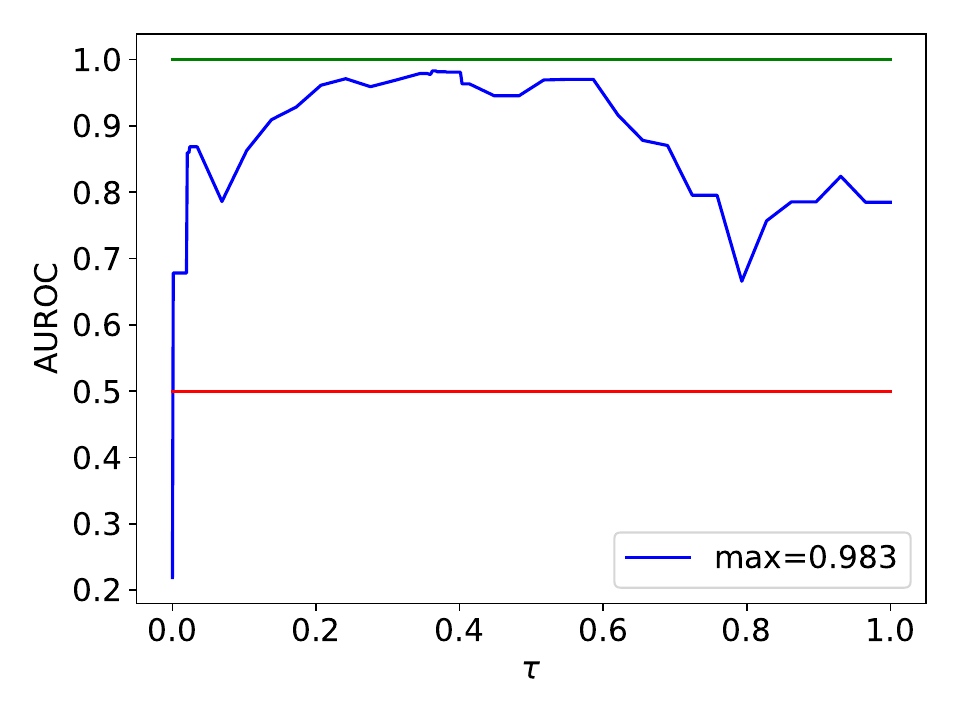}
    \end{subfigure}
    \hfill
    \begin{subfigure}{0.180\linewidth}
    \includegraphics[width=\linewidth]{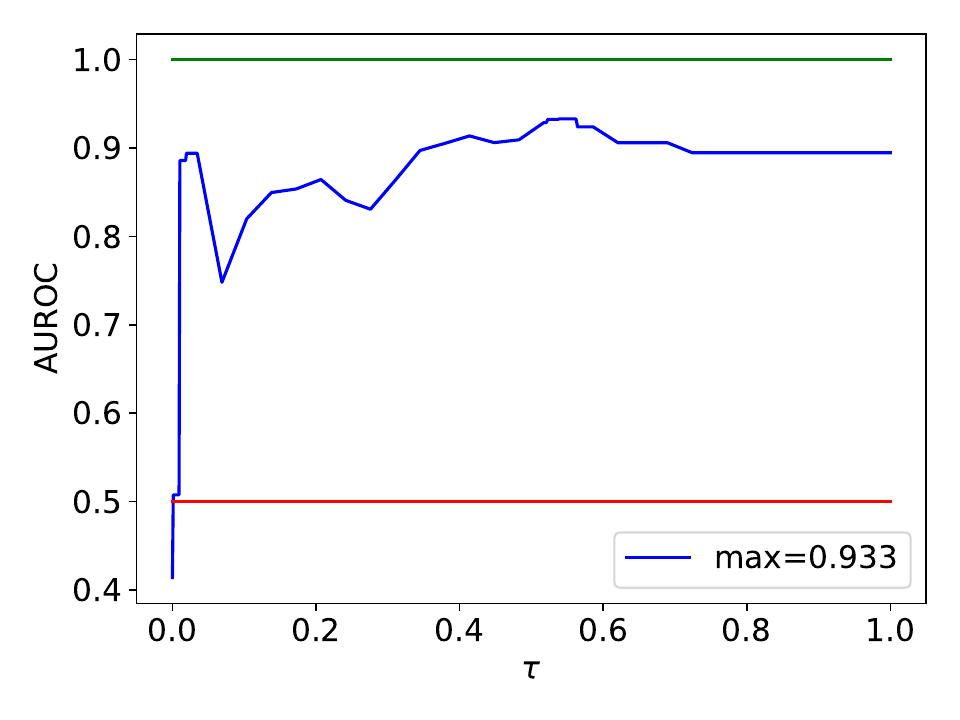}
    \end{subfigure}
    \hfill
    \begin{subfigure}{0.180\linewidth}
    \includegraphics[width=\linewidth]{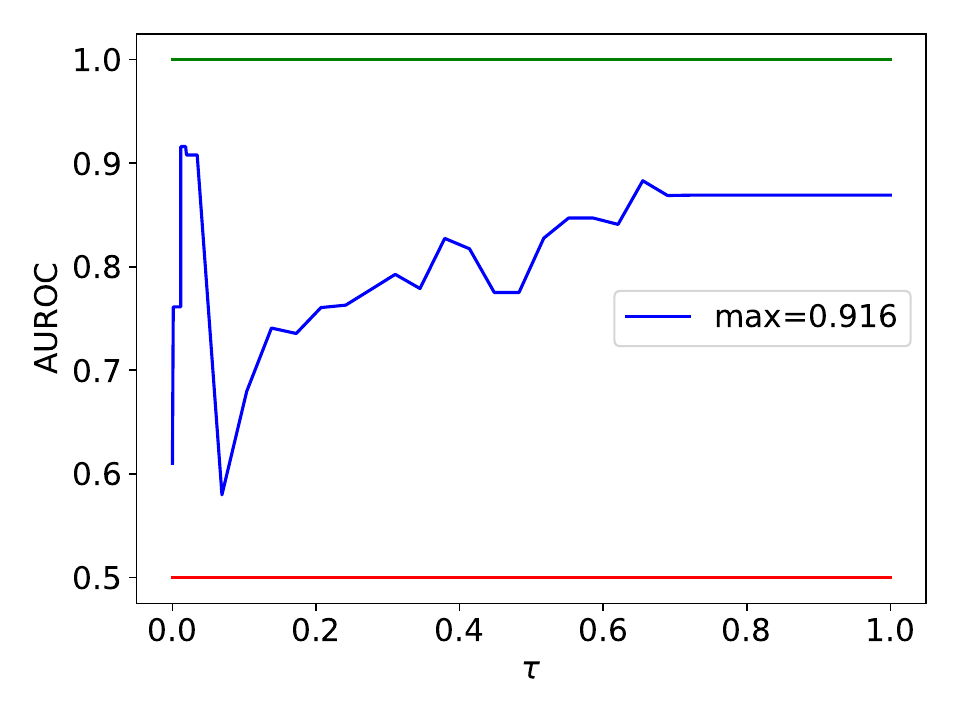}
    \end{subfigure}
    \hfill
    \begin{subfigure}{0.180\linewidth}
    \includegraphics[width=\linewidth]{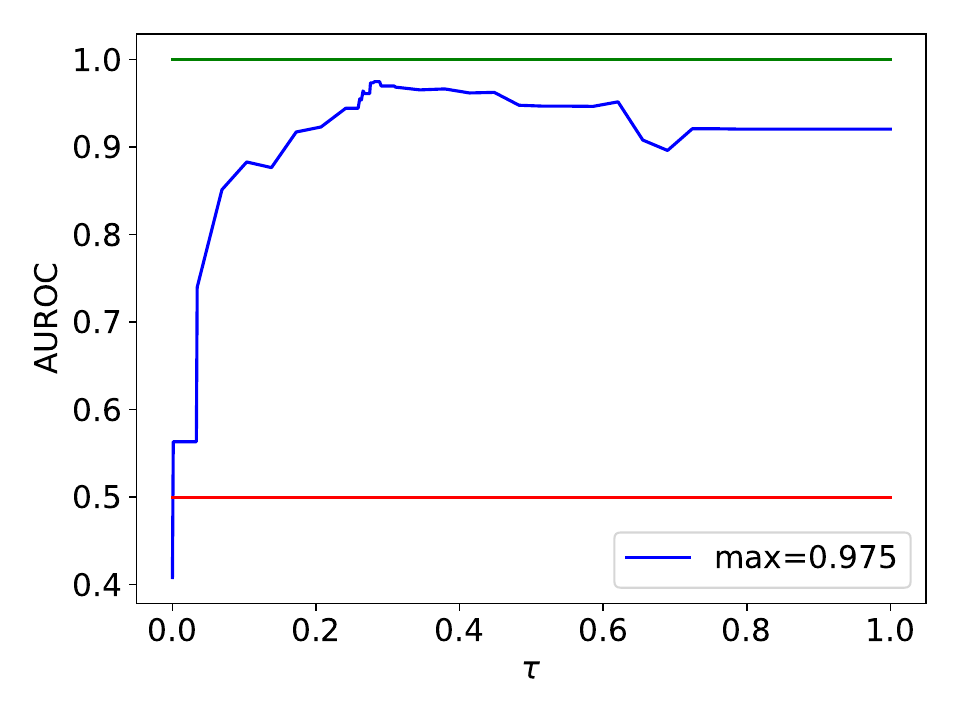}
    \end{subfigure}
    \hfill
    \begin{subfigure}{0.180\linewidth}
    \includegraphics[width=\linewidth]{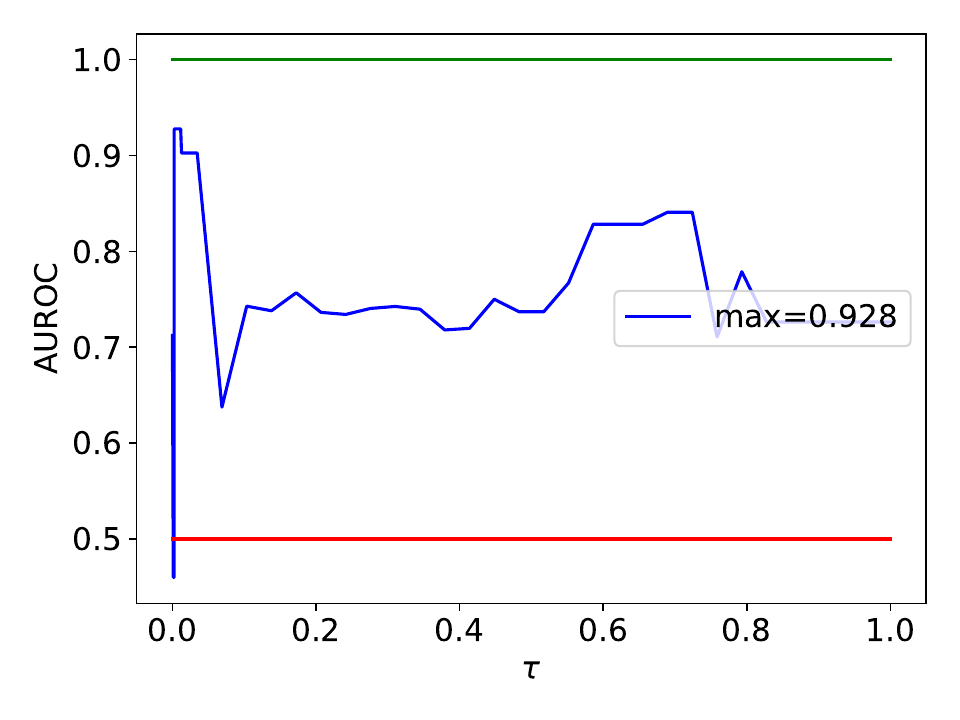}
    \end{subfigure}
    \hfill
    \begin{subfigure}{0.180\linewidth}
    \includegraphics[width=\linewidth]{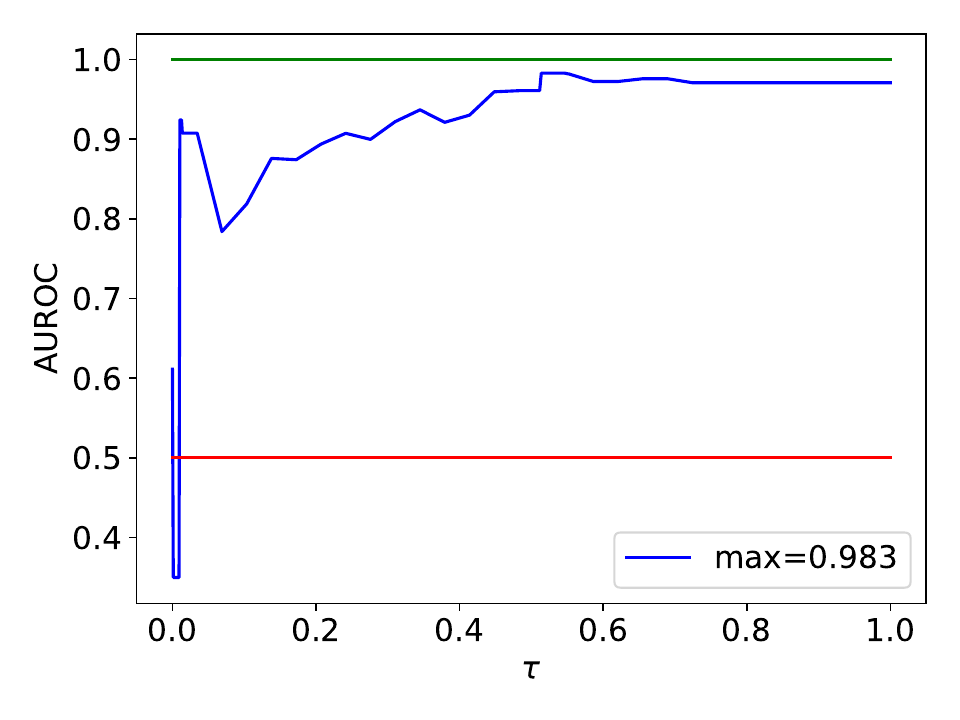}
    \end{subfigure}
    \hfill
    \begin{subfigure}{0.180\linewidth}
    \includegraphics[width=\linewidth]{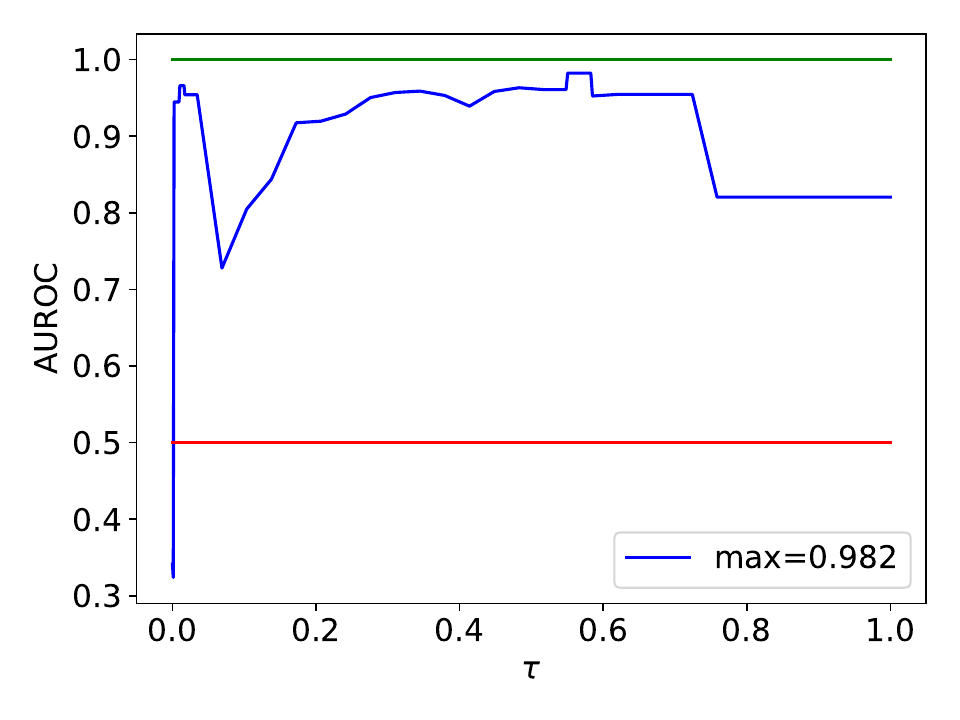}
    \end{subfigure}
    \hfill
    \begin{subfigure}{0.180\linewidth}
    \includegraphics[width=\linewidth]{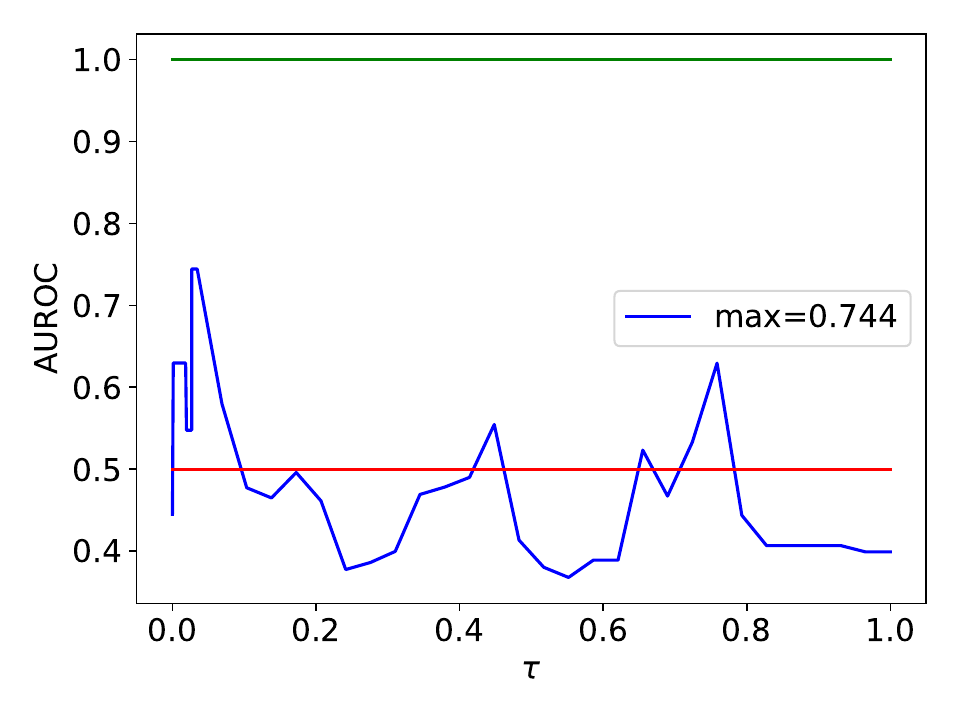}
    \end{subfigure}
    \hfill
    \begin{subfigure}{0.180\linewidth}
    \includegraphics[width=\linewidth]{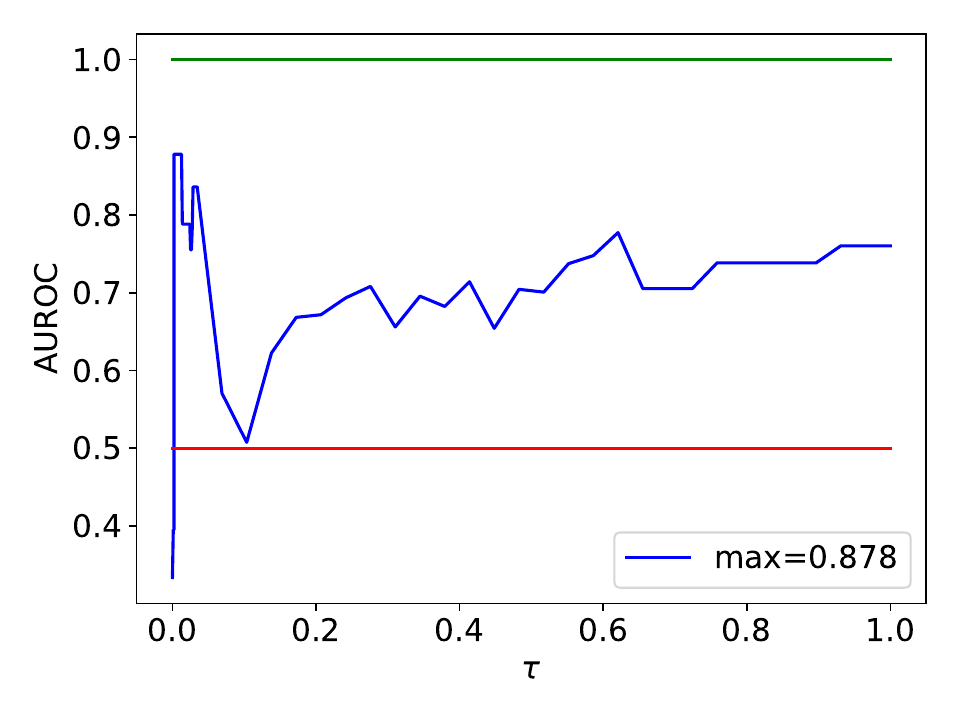}
    \end{subfigure}
    \hfill
    \caption{AUROC results obtained with ACGCTNAD for different $\tau$ values considering each type of digit as normal and all others as anomalous. In green we mark the maximum possible value ($1$) and in red the value $0.5$.}
    \label{fig:AUROC_vs_tau_Digits}
\end{figure}

\begin{figure}
    \centering
    \includegraphics[width=0.5\linewidth]{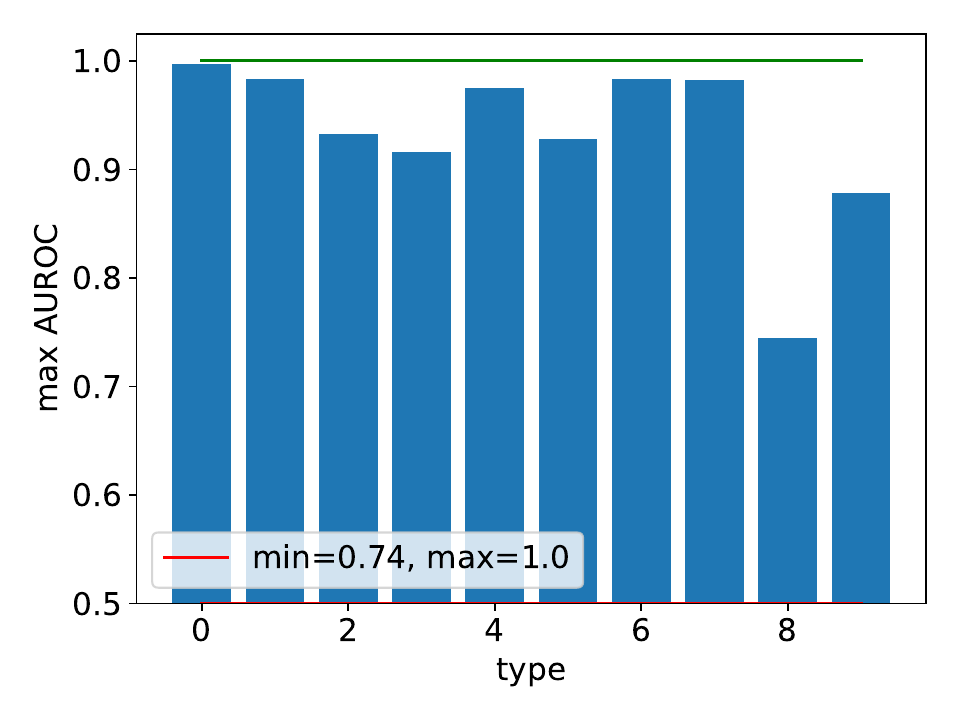}
    \caption{Maximum AUROC achieved with ACGCTNAD for each type of digit. In green we mark the maximum possible value ($1$) and in red the minimum possible value ($0.5$).}
    \label{fig:max_AUROC}
\end{figure}

In Fig. \ref{fig:AUROC_vs_tau_Digits} we can see the sorted results obtained with the ACGCTNAD method for each digit as normal, and in Fig. \ref{fig:max_AUROC} we can see the maximum AUROC obtained for each digit. In all cases we can see that the performance tends to improve for high compression values in the cases where it can be well distinguished, having in most occasions a performance peak for very low and concrete compression values, which may seem counter-intuitive. One explanation we can give for this phenomenon is that, for certain low values of compression, the representation deletes much of the structure of the anomalous data, but for higher values it starts to delete some of the structure of the normal data.

We can also observe that all digits yield a maximum AUROC of at least $0.74$, being at best $0.997$, with 0, 1, 4, 6 and 7 being the best, and 8 and 9 the worst.

\begin{figure}
    \centering
    \begin{subfigure}{0.180\linewidth}
    \includegraphics[width=\linewidth]{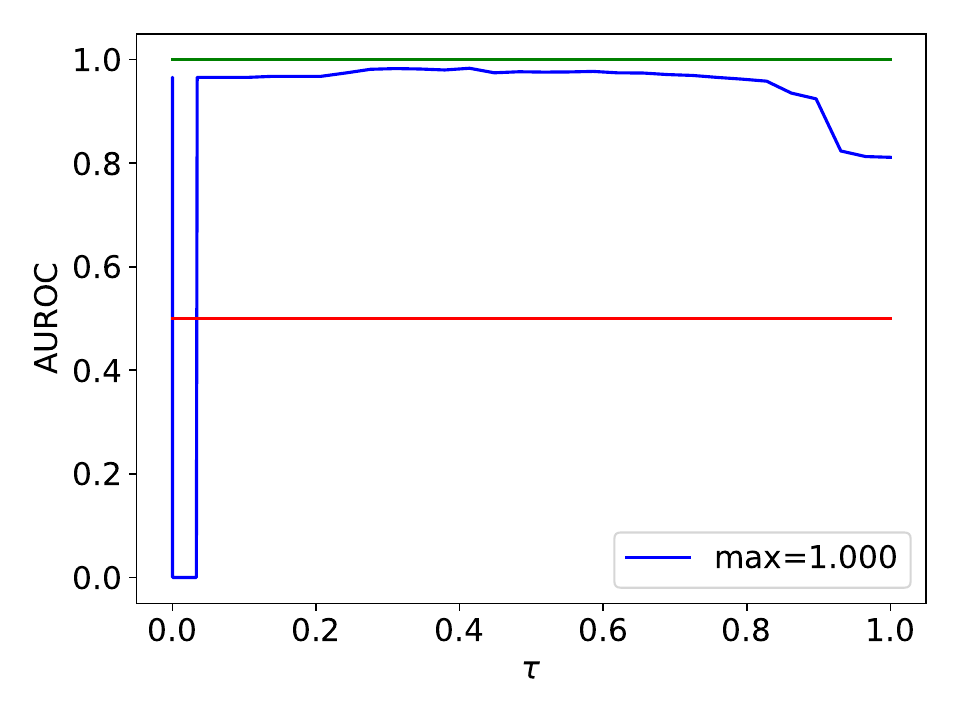}
    \end{subfigure}
    \hfill
    \begin{subfigure}{0.180\linewidth}
    \includegraphics[width=\linewidth]{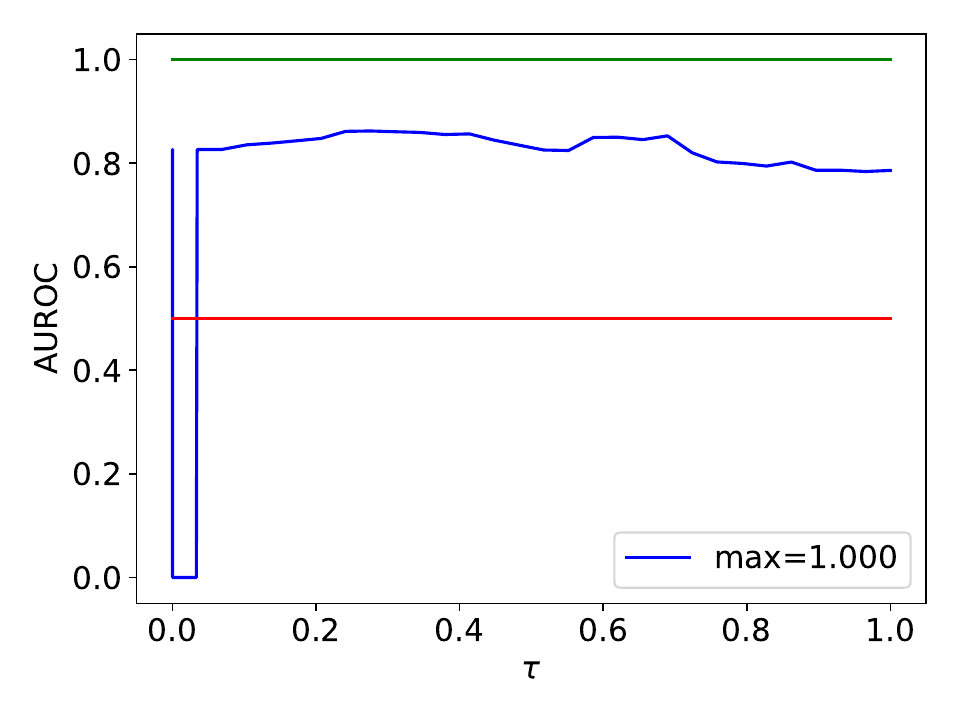}
    \end{subfigure}
    \hfill
    \begin{subfigure}{0.180\linewidth}
    \includegraphics[width=\linewidth]{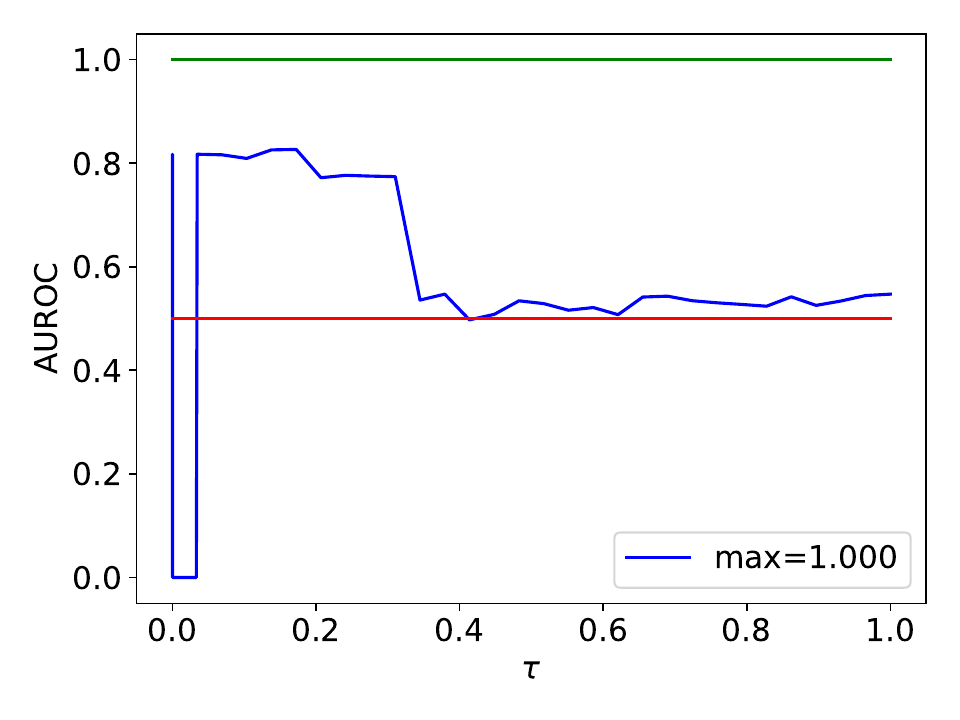}
    \end{subfigure}
    \hfill
    \begin{subfigure}{0.180\linewidth}
    \includegraphics[width=\linewidth]{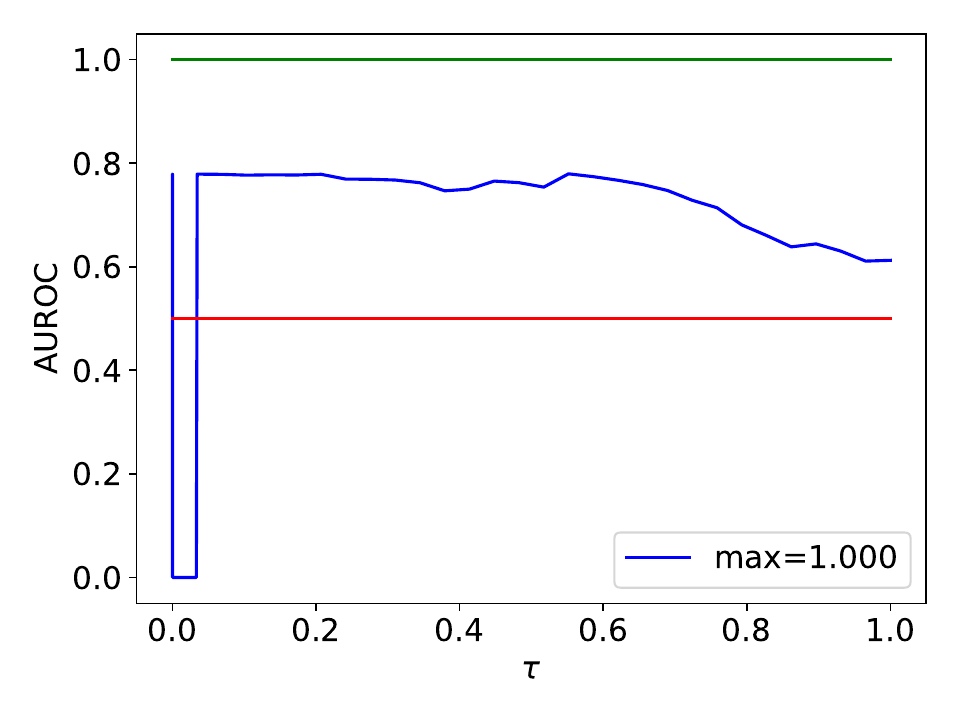}
    \end{subfigure}
    \hfill
    \begin{subfigure}{0.180\linewidth}
    \includegraphics[width=\linewidth]{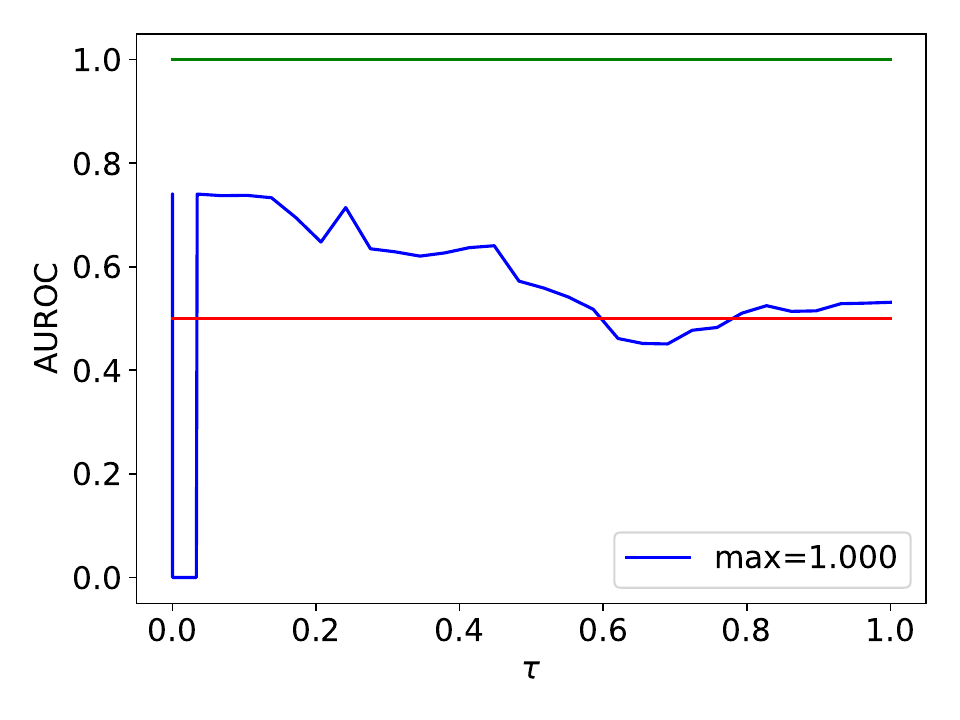}
    \end{subfigure}
    \hfill
    \begin{subfigure}{0.180\linewidth}
    \includegraphics[width=\linewidth]{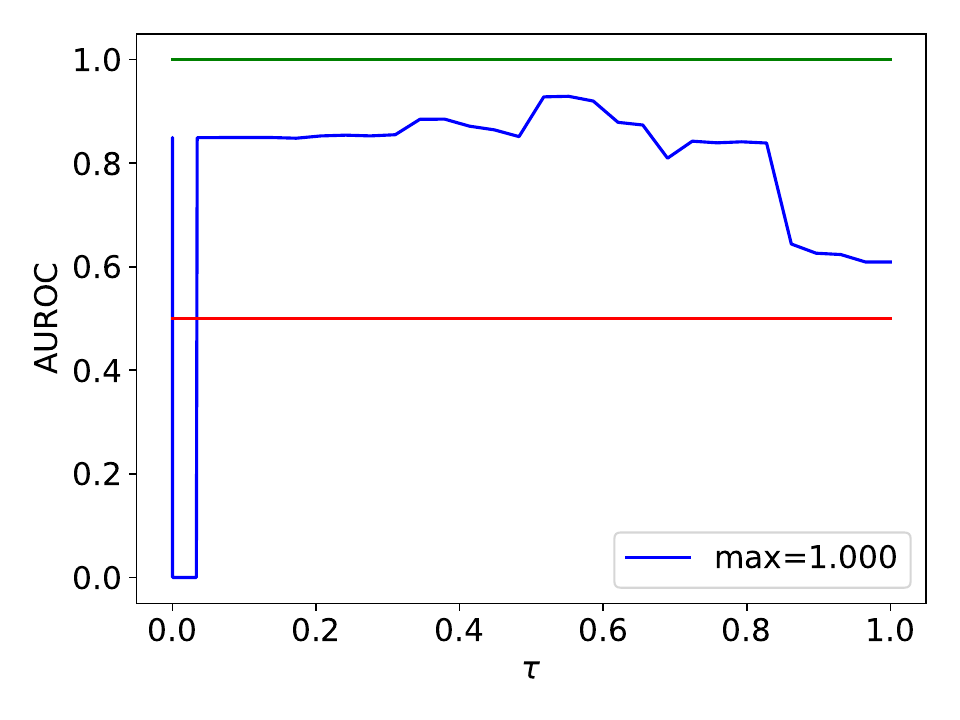}
    \end{subfigure}
    \hfill
    \begin{subfigure}{0.180\linewidth}
    \includegraphics[width=\linewidth]{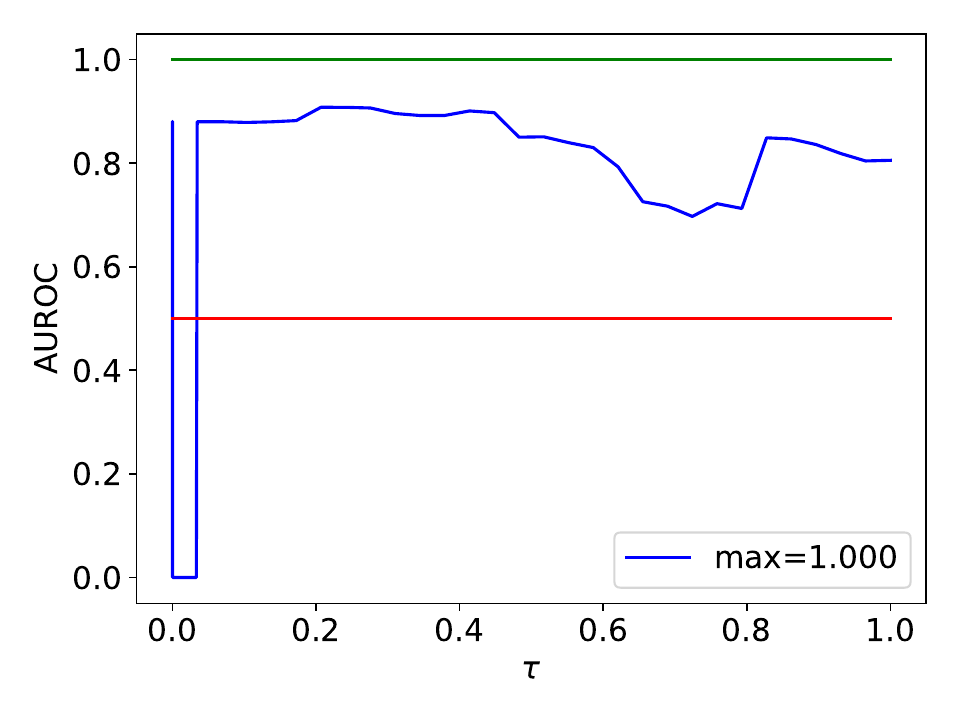}
    \end{subfigure}
    \hfill
    \begin{subfigure}{0.180\linewidth}
    \includegraphics[width=\linewidth]{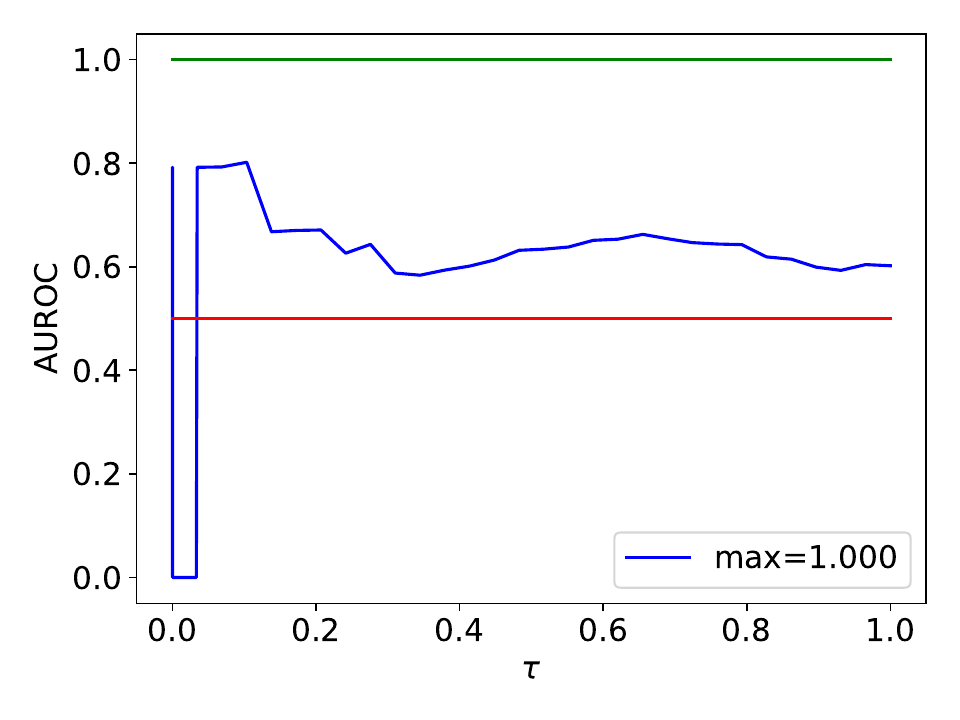}
    \end{subfigure}
    \hfill
    \begin{subfigure}{0.180\linewidth}
    \includegraphics[width=\linewidth]{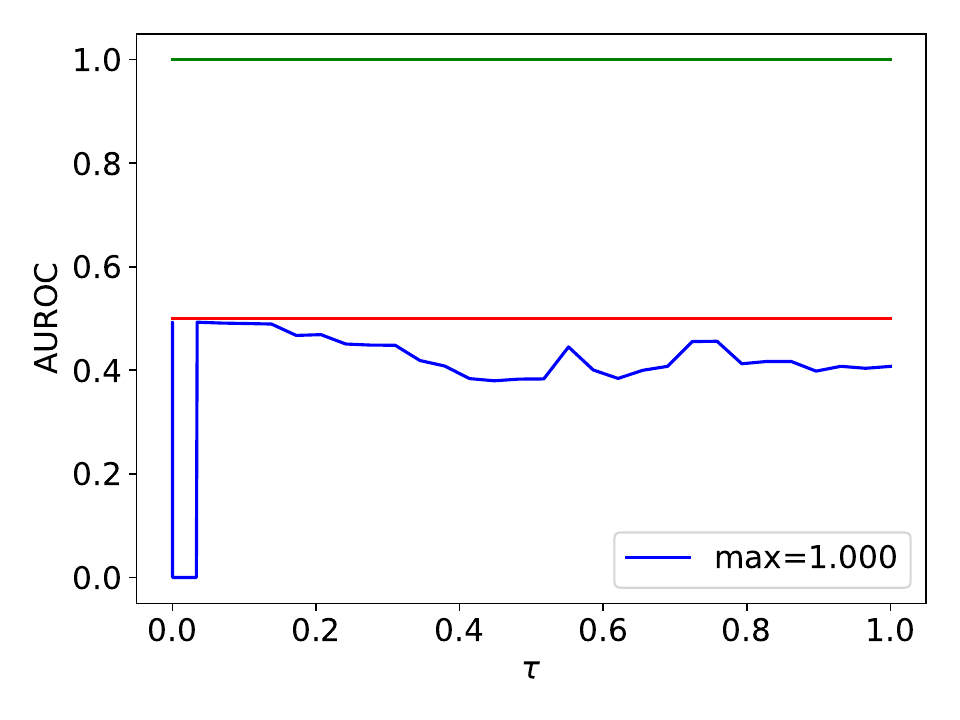}
    \end{subfigure}
    \hfill
    \begin{subfigure}{0.180\linewidth}
    \includegraphics[width=\linewidth]{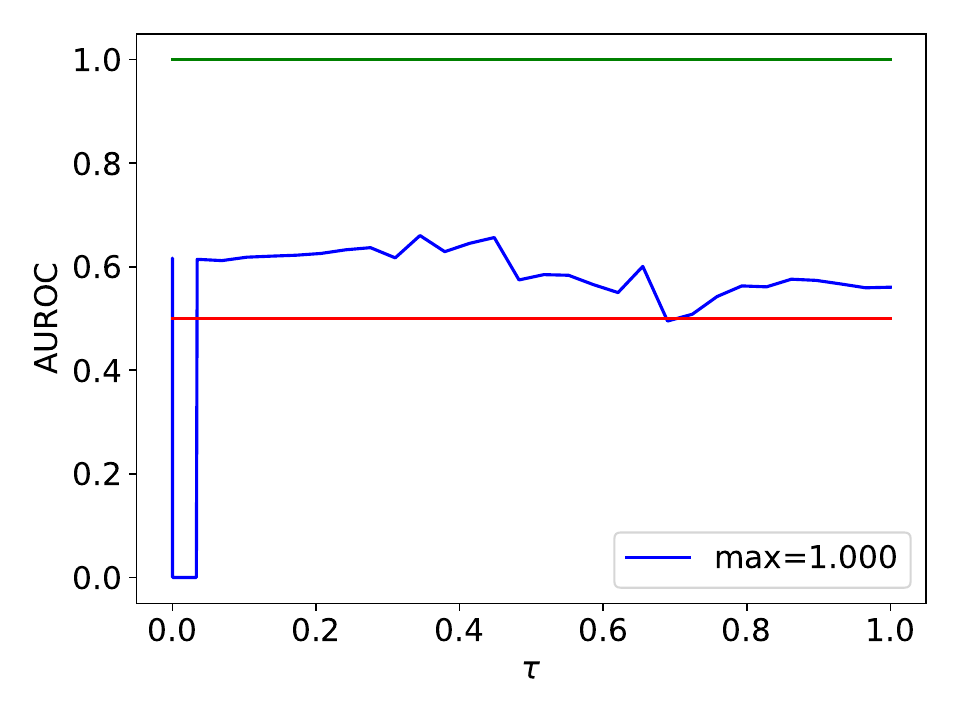}
    \end{subfigure}
    \hfill
    \caption{AUROC results obtained with ACLCTNAD for different $\tau$ values considering each type of digit as normal and all others as anomalous. In green we mark the maximum possible value ($1$) and in red the value $0.5$.}
    \label{fig:AUROC_vs_tau_Digits_local}
\end{figure}

\begin{figure}
    \centering
    \includegraphics[width=0.5\linewidth]{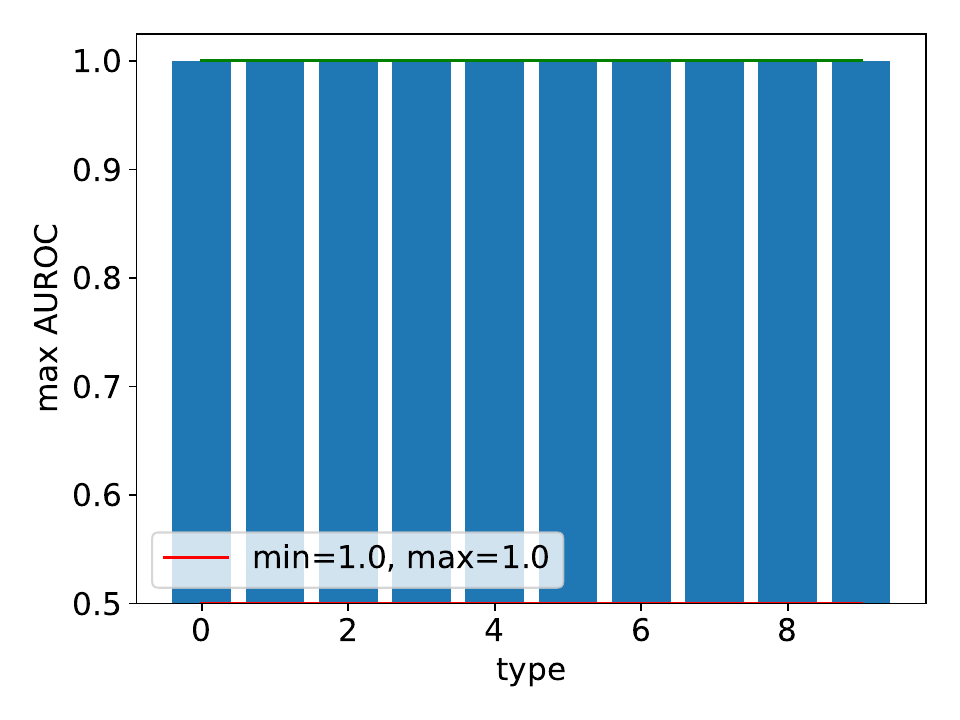}
    \caption{Maximum AUROC achieved with ACLCTNAD for each type of digit. In green we mark the maximum possible value ($1$) and in red the minimum possible value ($0.5$).}
    \label{fig:max_AUROC_local}
\end{figure}

In Fig. \ref{fig:AUROC_vs_tau_Digits_local} we can see the sorted results obtained with the ACLCTNAD method for each digit as normal, and in Fig. \ref{fig:max_AUROC_local} we can see the maximum AUROC obtained for each digit, in the same way as for the global method. In this case we see an exceptionally strange situation. While most of the performance distribution is stable, without being good in many of the cases, in all cases there is a small region of low compression in which the AUROC becomes $0$, meaning that the score orientation is reversed with respect to the anomaly labels. If one inverts the score after inspecting the test labels, the inverted AUROC becomes $1$ in that regime. This observation is diagnostically informative because it shows that the score still contains discriminative information, but such a post-hoc inversion is supervised and should not be reported as an unsupervised test result unless the inversion rule has been fixed independently, for example from prior knowledge, from the training protocol, or from a separate validation set. This behavior can be seen as an extreme case of the orientation phenomenon already suggested by the global method. However, the extreme slowness of the method makes us not recommend it except in specific cases where such accuracy is desired at any cost.

If we compare the performance of both methods, we see that method ACLCTNAD provides better results for all of the cases, but the ACGCTNAD has exceptionally good results, especially for certain classes, being 50 times faster. This makes it clear that the global method is more effective for this type of cases, but the local method is still good for certain cases. 

We can perform the same tests without applying any scaler to the data, to see if this transformation affects the performance of the solver. Due to the previous result and the modifications it introduces, we will perform this test on the global method, since it would have no noticeable effect on the local method.

\begin{figure}
    \centering
    \begin{subfigure}{0.180\linewidth}
    \includegraphics[width=\linewidth]{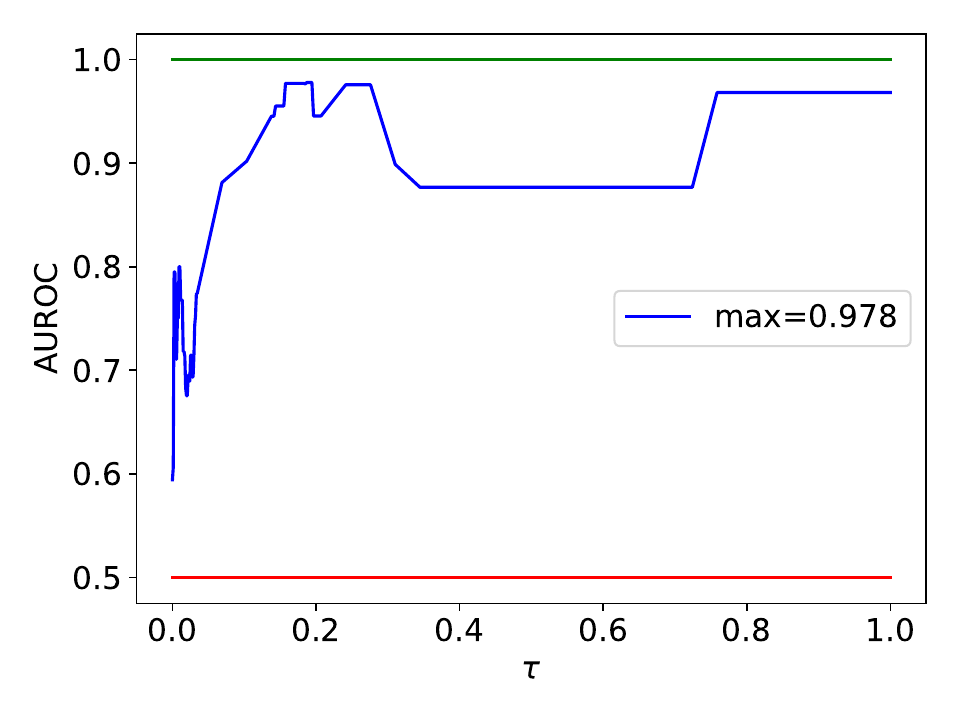}
    \end{subfigure}
    \hfill
    \begin{subfigure}{0.180\linewidth}
    \includegraphics[width=\linewidth]{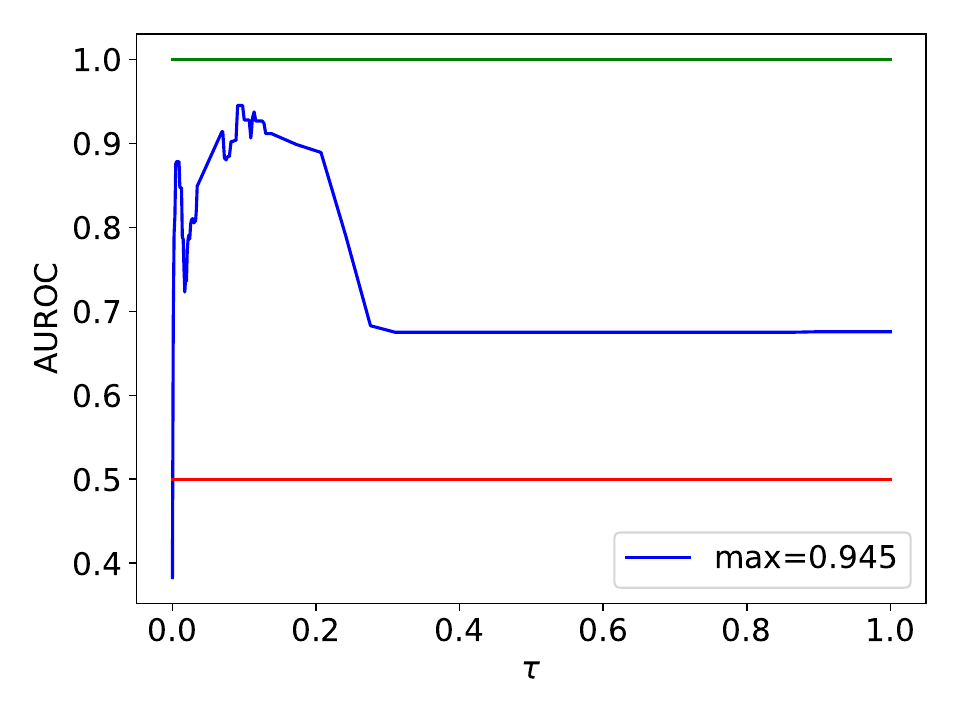}
    \end{subfigure}
    \hfill
    \begin{subfigure}{0.180\linewidth}
    \includegraphics[width=\linewidth]{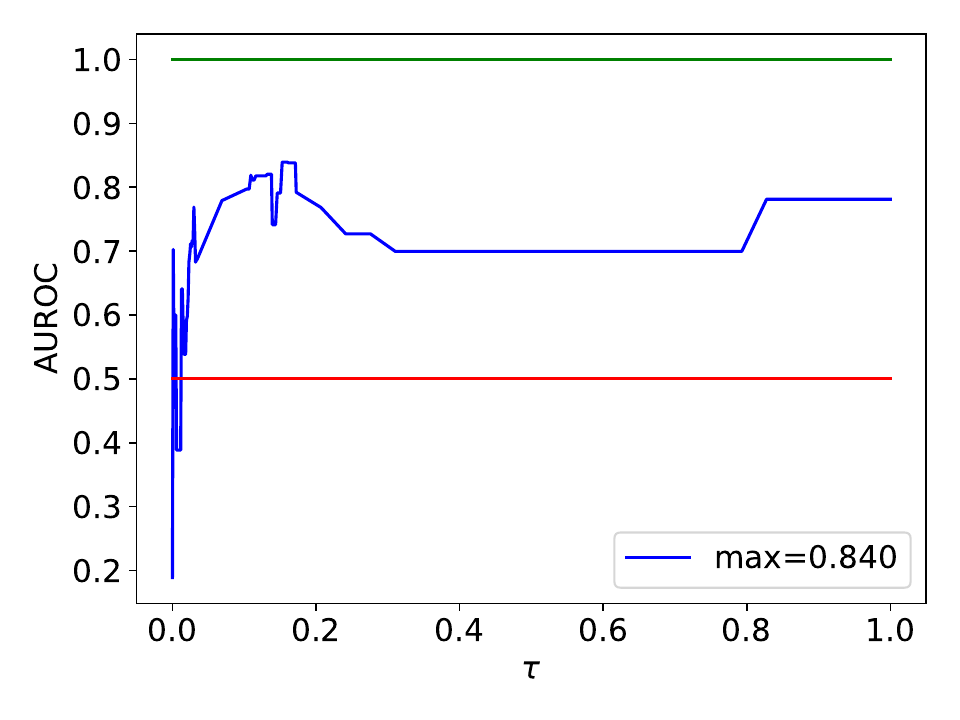}
    \end{subfigure}
    \hfill
    \begin{subfigure}{0.180\linewidth}
    \includegraphics[width=\linewidth]{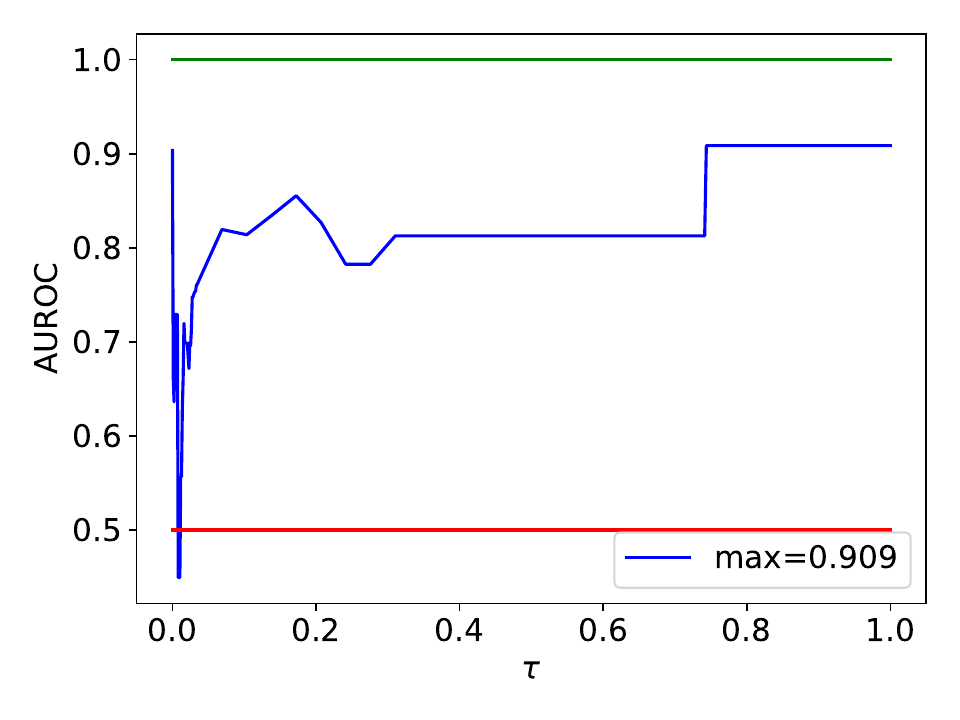}
    \end{subfigure}
    \hfill
    \begin{subfigure}{0.180\linewidth}
    \includegraphics[width=\linewidth]{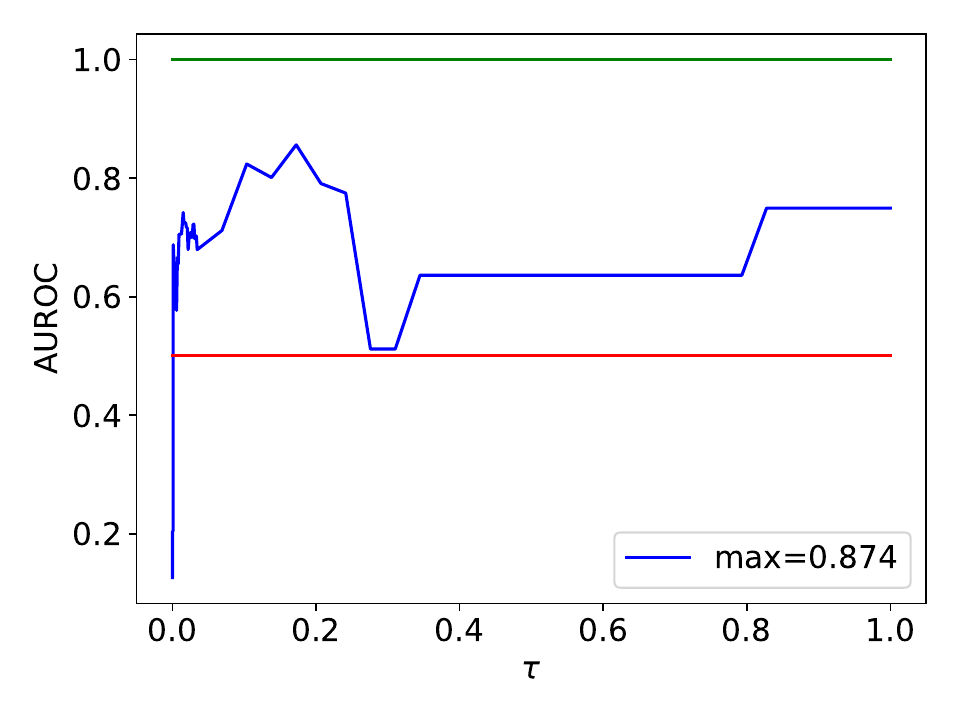}
    \end{subfigure}
    \hfill
    \begin{subfigure}{0.180\linewidth}
    \includegraphics[width=\linewidth]{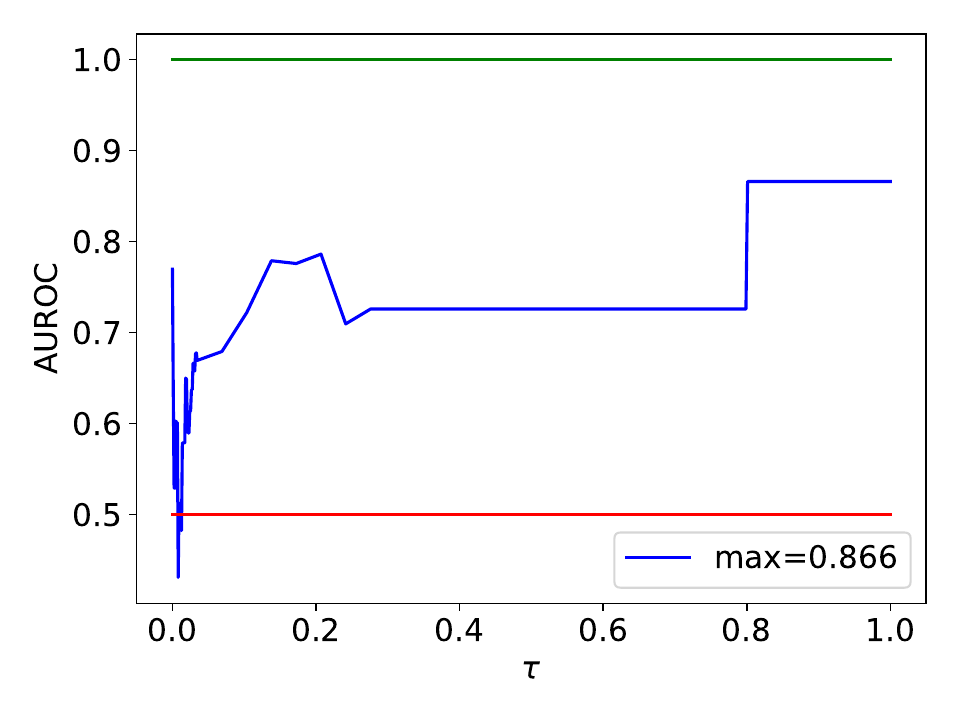}
    \end{subfigure}
    \hfill
    \begin{subfigure}{0.180\linewidth}
    \includegraphics[width=\linewidth]{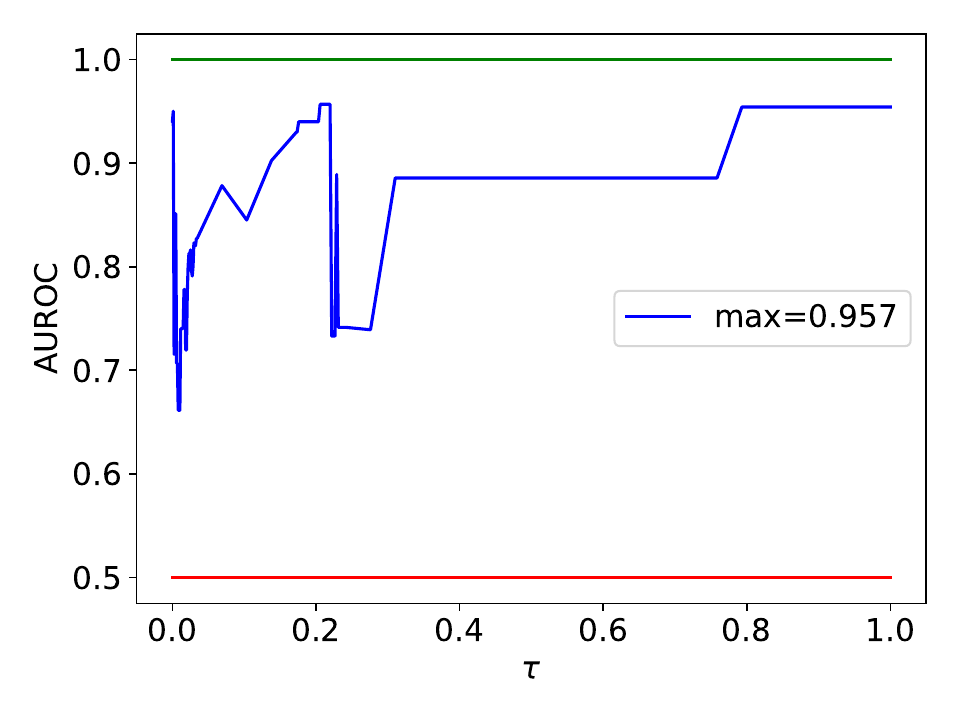}
    \end{subfigure}
    \hfill
    \begin{subfigure}{0.180\linewidth}
    \includegraphics[width=\linewidth]{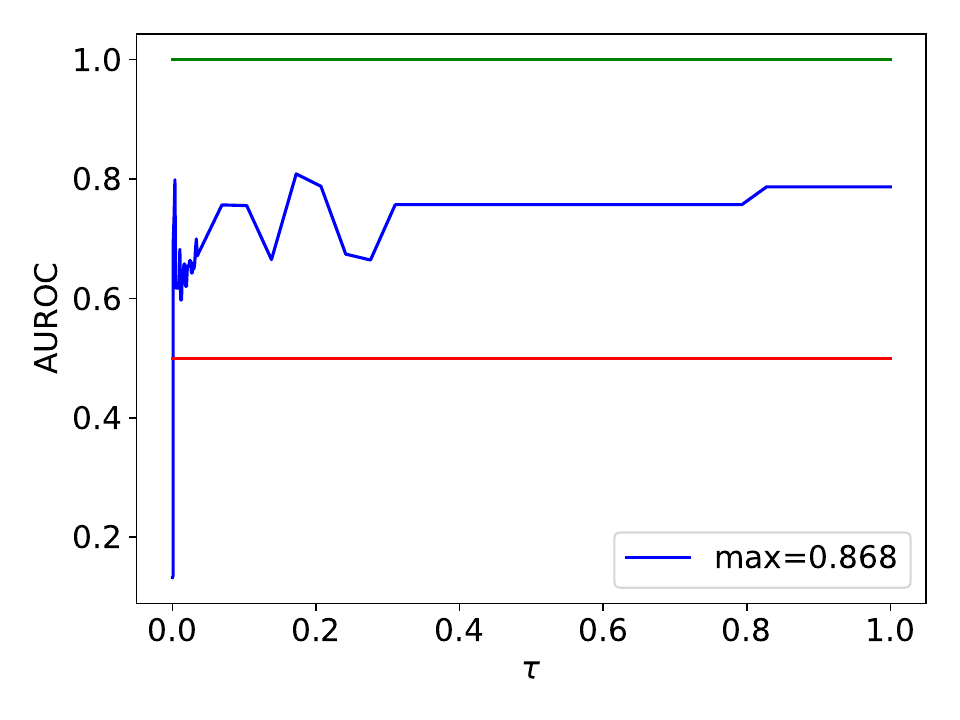}
    \end{subfigure}
    \hfill
    \begin{subfigure}{0.180\linewidth}
    \includegraphics[width=\linewidth]{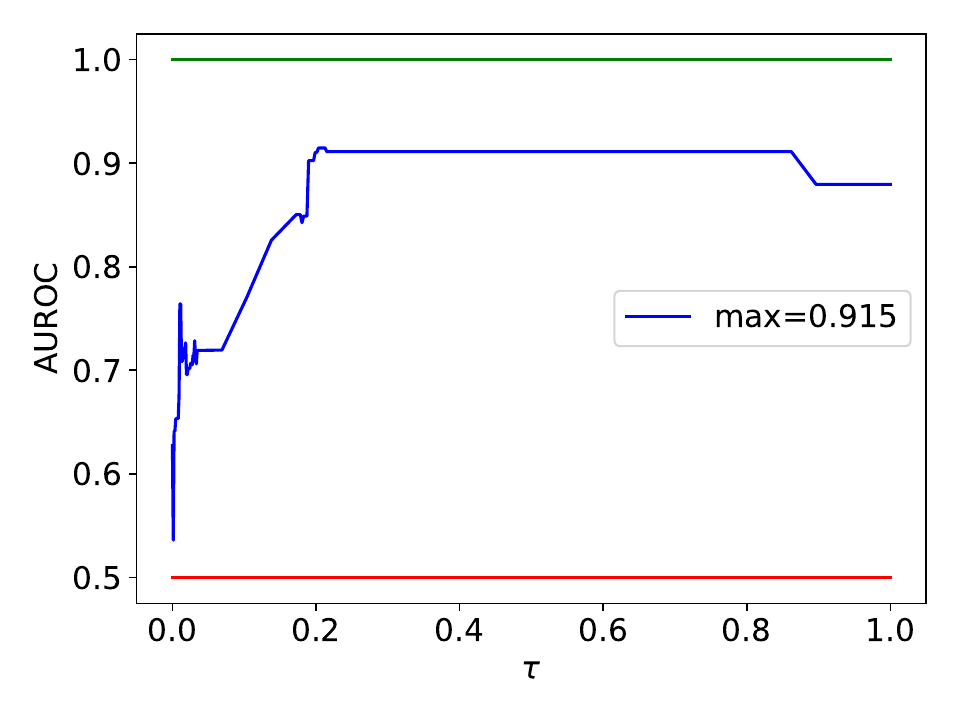}
    \end{subfigure}
    \hfill
    \begin{subfigure}{0.180\linewidth}
    \includegraphics[width=\linewidth]{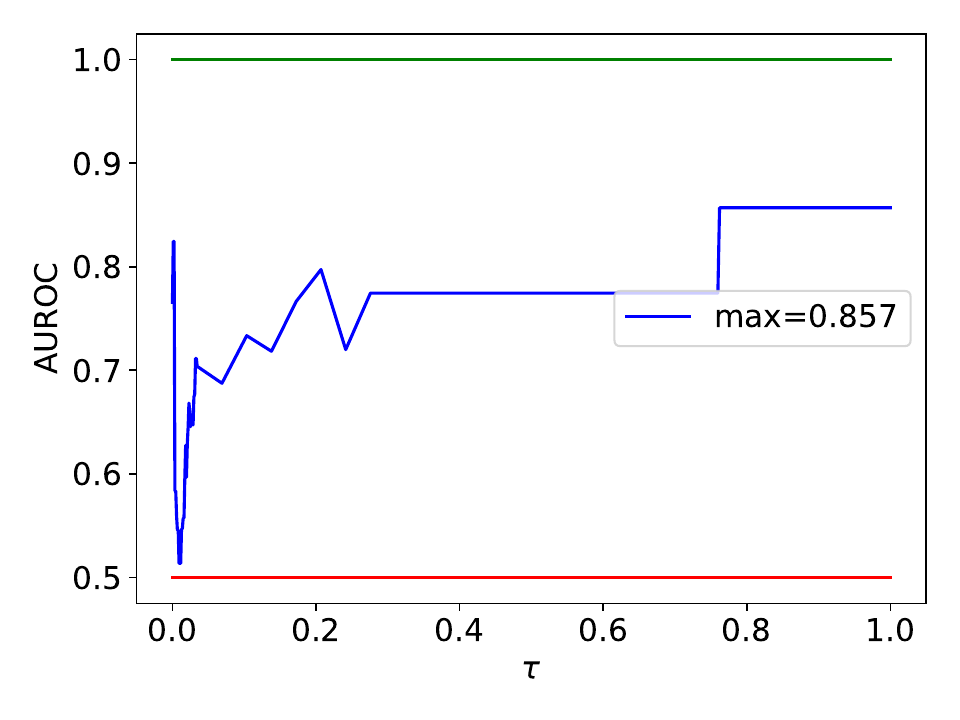}
    \end{subfigure}
    \hfill
    \caption{AUROC results obtained with ACGCTNAD without scaling for different $\tau$ values considering each type of digit as normal and all others as anomalous. In green we mark the maximum possible value ($1$) and in red the value $0.5$.}
    \label{fig:AUROC_vs_tau_Digits_non_scaler}
\end{figure}

\begin{figure}
    \centering
    \includegraphics[width=0.5\linewidth]{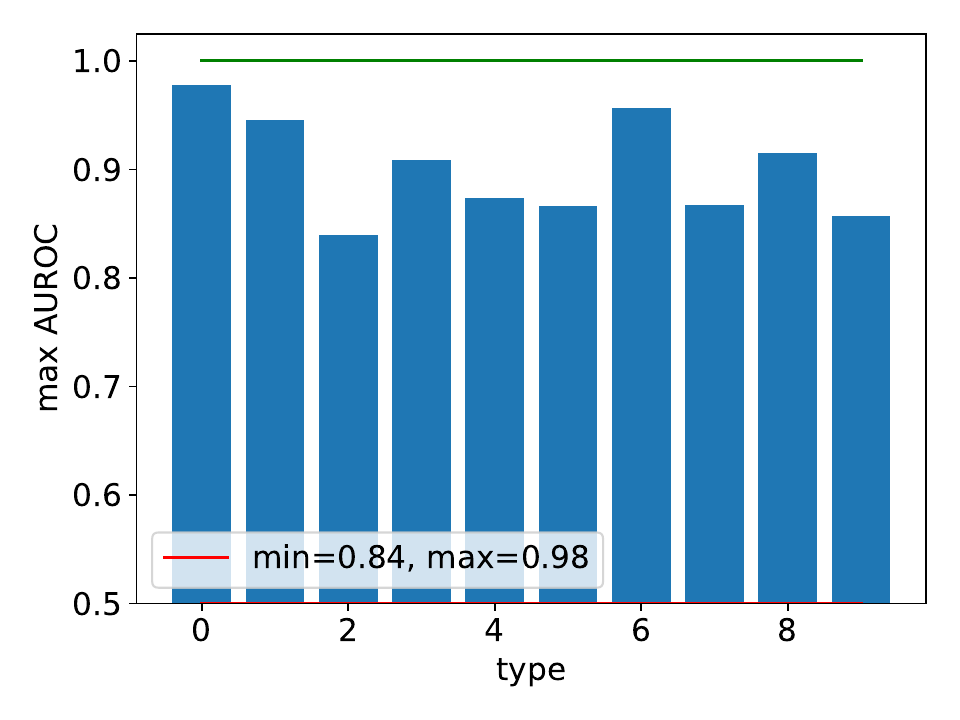}
    \caption{Maximum AUROC achieved with ACGCTNAD for each type of digit. In green we mark the maximum possible value ($1$) and in red the minimum possible value ($0.5$).}
    \label{fig:max_AUROC_non_scaler}
\end{figure}

In Fig. \ref{fig:AUROC_vs_tau_Digits_non_scaler} and \ref{fig:max_AUROC_non_scaler} we can see the sorted results obtained with the ACGCTNAD without scaling. In this case we see the same behavior profile as with the scaler case, but with a lower overall performance, except for the previously worst case, digit 8, which has gone from a maximum AUROC of $0.744$ to $0.915$.

We can repeat the same test for the Olivetti faces dataset, focusing on the global method. In this case we normalize the data with a standard scaler and use a shape of $\{2,2,2,2,2,2,2,2,2,2,2,2\}$. For each test, we will consider one of the face classes as the normal one and the rest as the anomalous ones. Due to the dataset size of 400 samples and 40 different classes, in each test we will work with about 8-9 normal data and the same number of anomalous ones.

\begin{figure}
    \centering
    \includegraphics[width=0.5\linewidth]{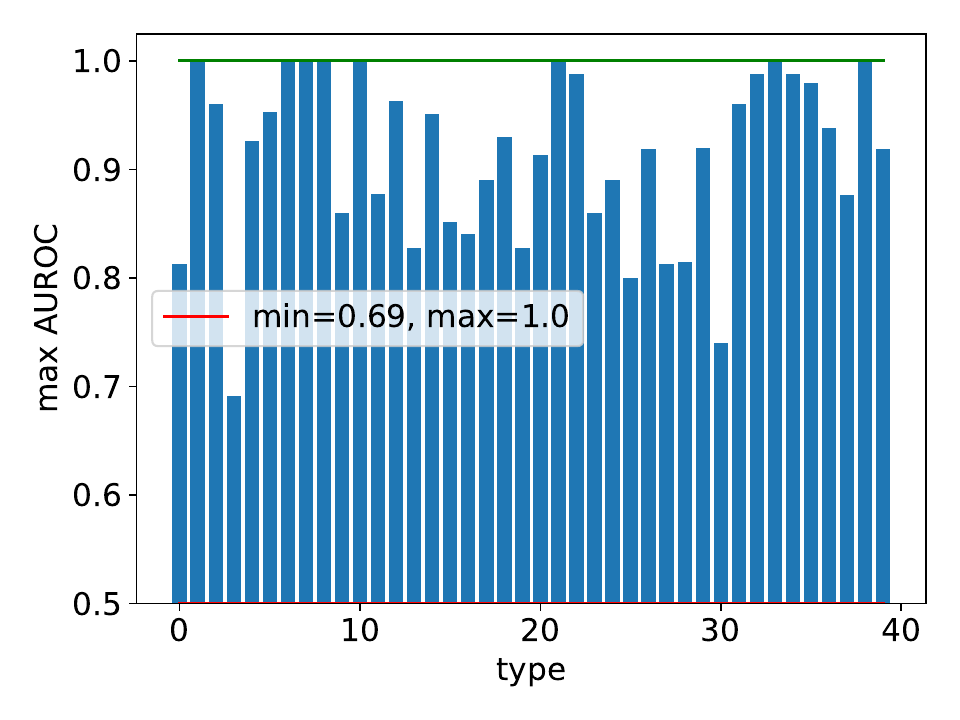}
    \caption{Maximum AUROC achieved with ACGCTNAD for each type of face with standard scaler. In green we mark the maximum possible value ($1$) and in red the minimum possible value ($0.5$).}
    \label{fig:olive}
\end{figure}

In Fig. \ref{fig:olive} we can see the result of the maximum AUROC obtained for each class considered as normal. We see that the ability to distinguish anomalies varies significantly between classes, with a minimum at $0.69$. This may be due to the small size of the sample available or because the method is not as well suited to this type of data.

We will now study the performance of the ACGCTNAD method for detecting cyber-attacks on a real cybersecurity dataset, in which there are records of requests made to a computer system, with 81 features per data, and composed of 4 groups: normal, brute force attack, scanning and slowloris. This dataset is internal and private, so we do not disclose further identifying information about it in the present paper. For this case, we will differentiate the normal data from the other 3 groups, which will count as anomalies. The shape chosen is $\{3,3,3,3\}$.

\begin{figure}
    \centering
    \begin{subfigure}{0.21\linewidth}
    \includegraphics[width=\linewidth]{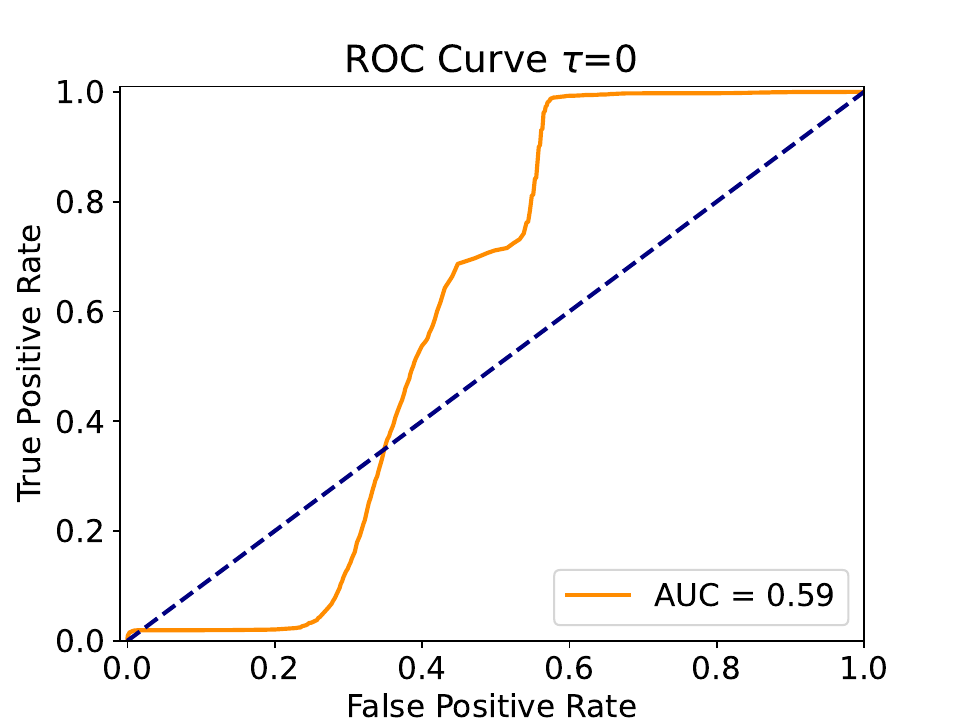}
    \end{subfigure}
    \hfill
    \begin{subfigure}{0.21\linewidth}
    \includegraphics[width=\linewidth]{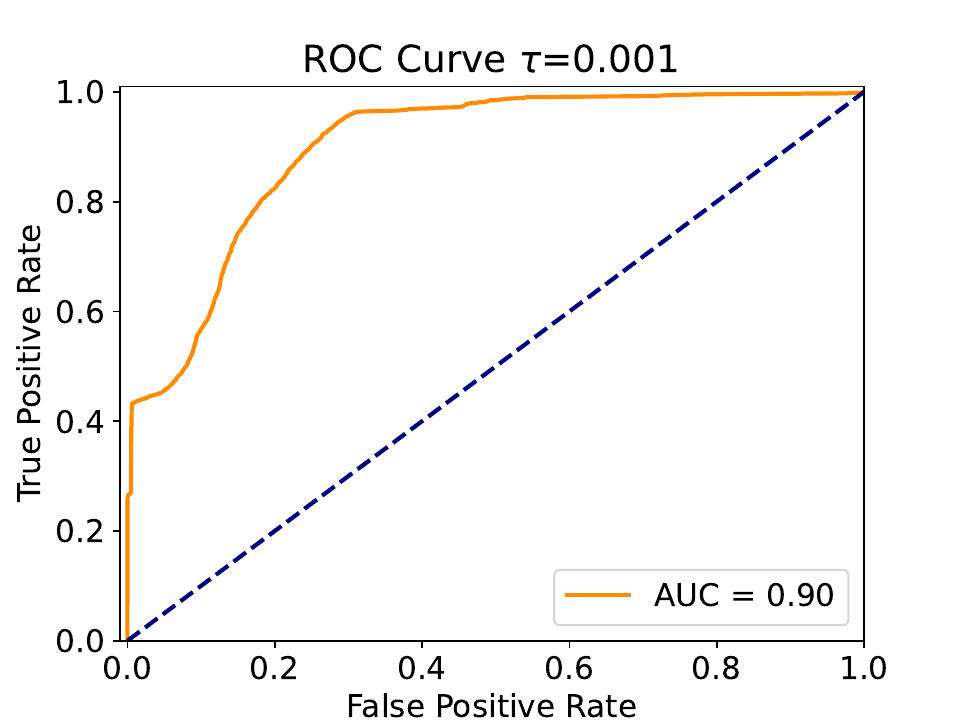}
    \end{subfigure}
    \hfill
    \begin{subfigure}{0.21\linewidth}
    \includegraphics[width=\linewidth]{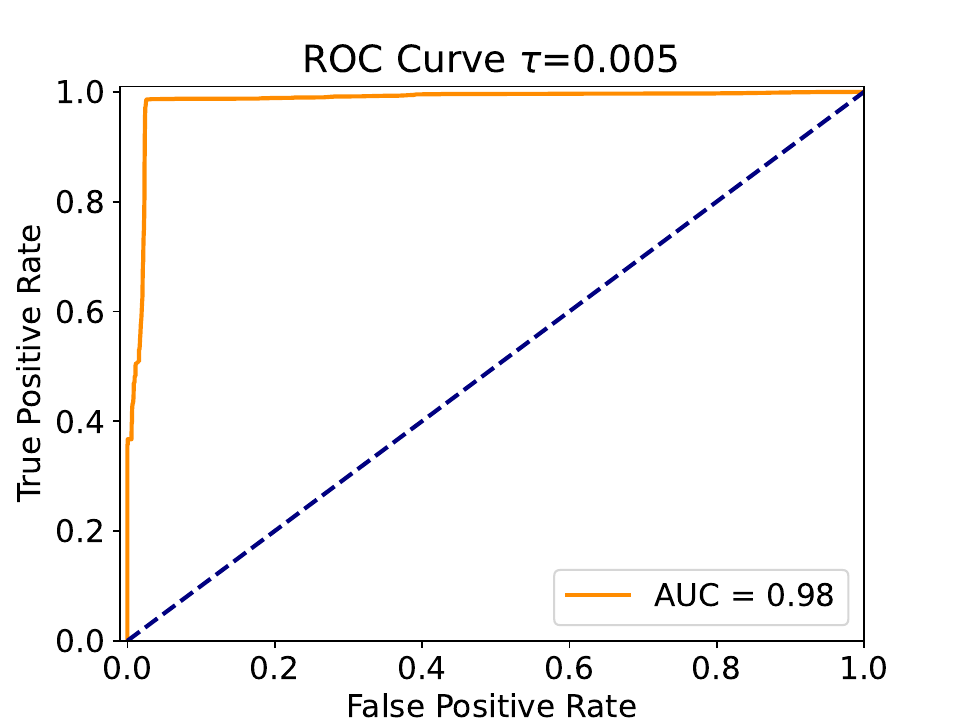}
    \end{subfigure}
    \hfill
    \begin{subfigure}{0.21\linewidth}
    \includegraphics[width=\linewidth]{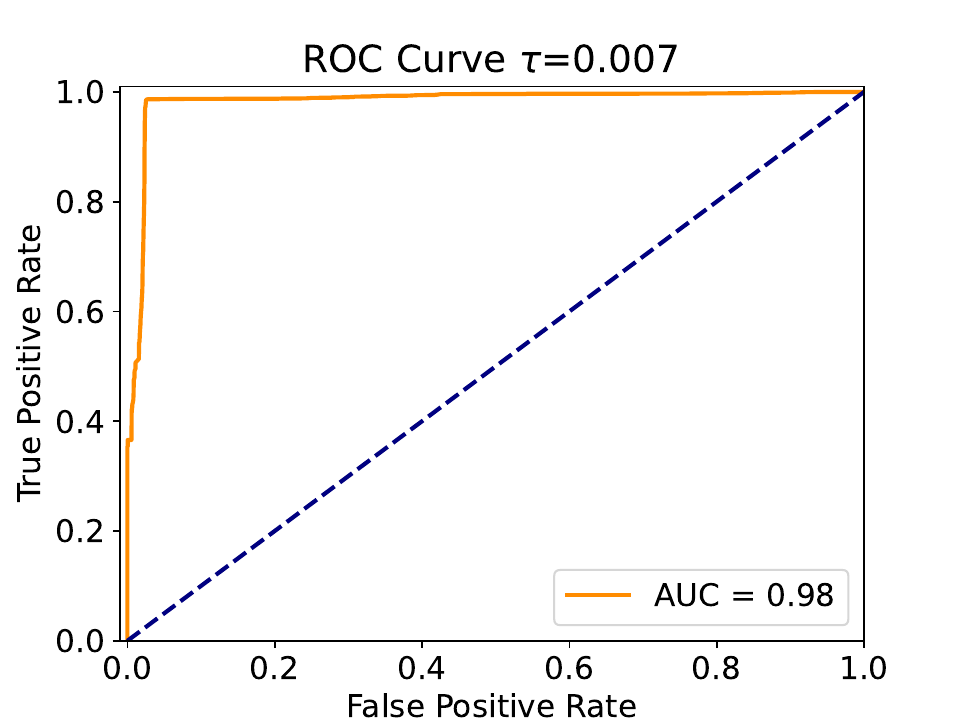}
    \end{subfigure}
    \hfill
    \begin{subfigure}{0.21\linewidth}
    \includegraphics[width=\linewidth]{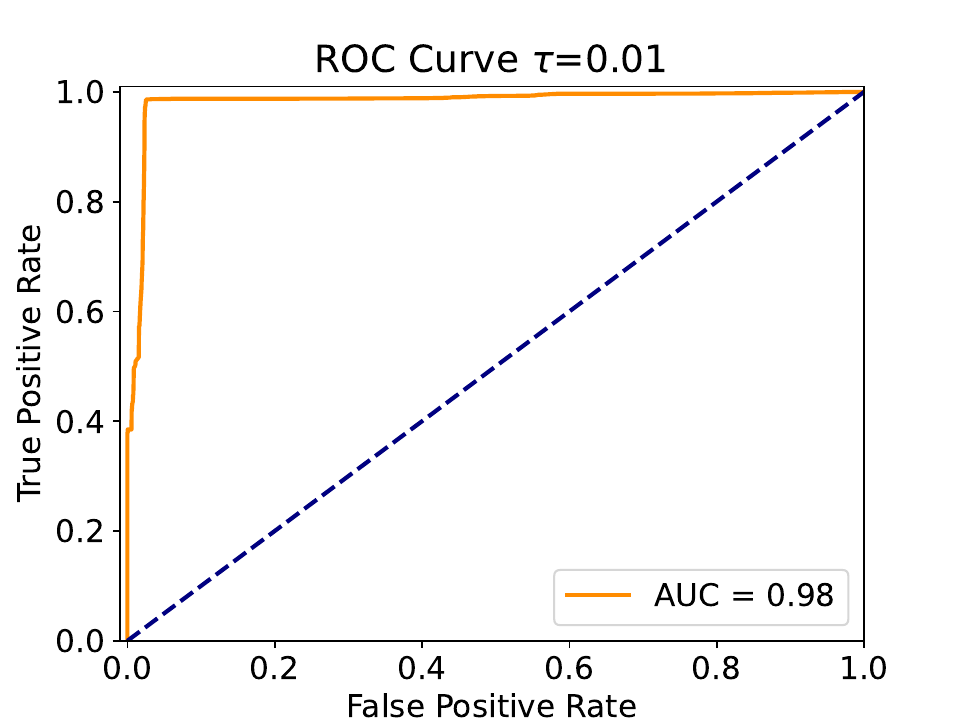}
    \end{subfigure}
    \hfill
    \begin{subfigure}{0.21\linewidth}
    \includegraphics[width=\linewidth]{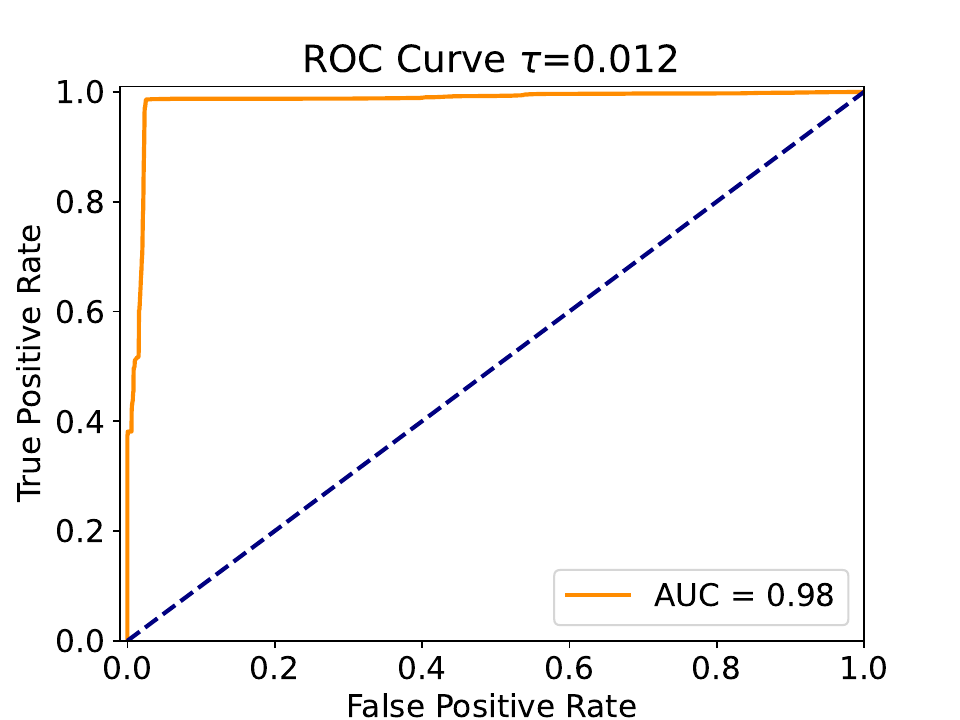}
    \end{subfigure}
    \hfill
    \begin{subfigure}{0.21\linewidth}
    \includegraphics[width=\linewidth]{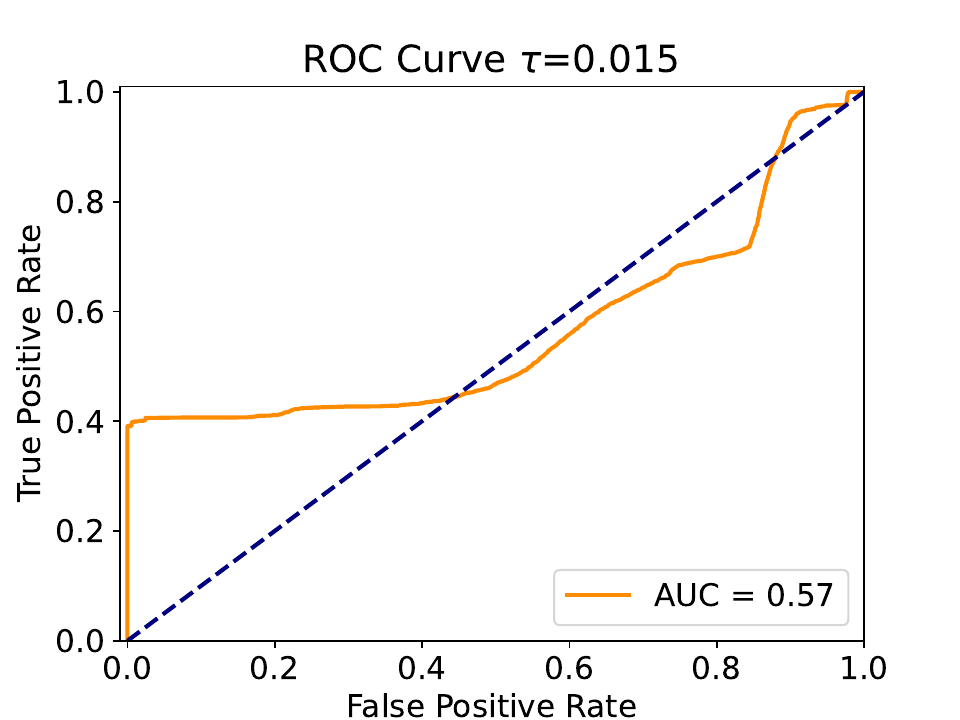}
    \end{subfigure}
    \hfill
    \begin{subfigure}{0.21\linewidth}
    \includegraphics[width=\linewidth]{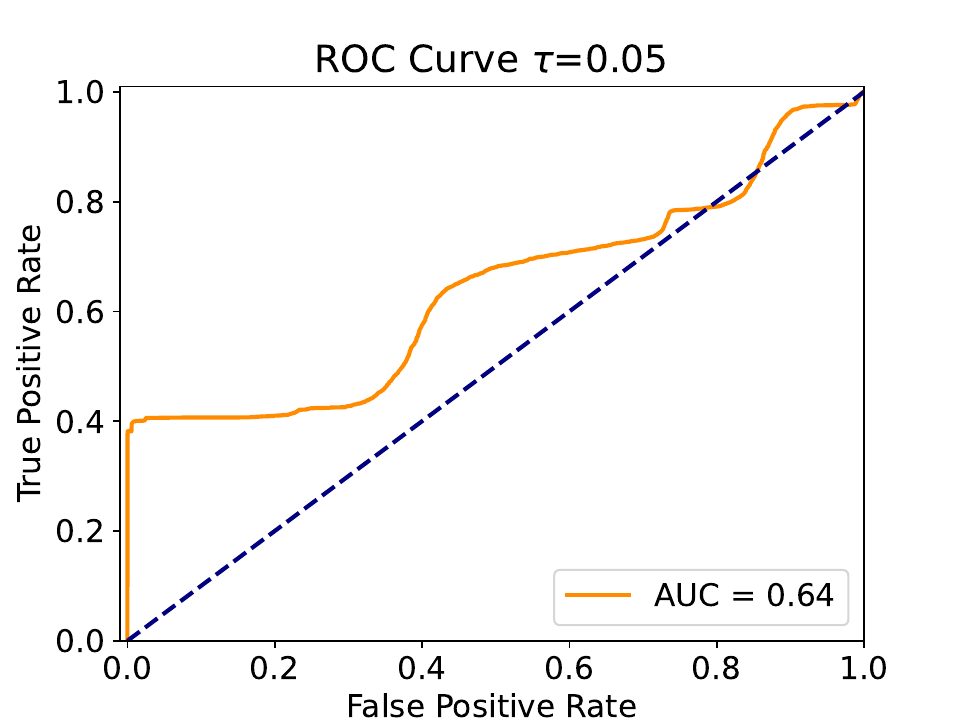}
    \end{subfigure}
    \hfill
    \begin{subfigure}{0.21\linewidth}
    \includegraphics[width=\linewidth]{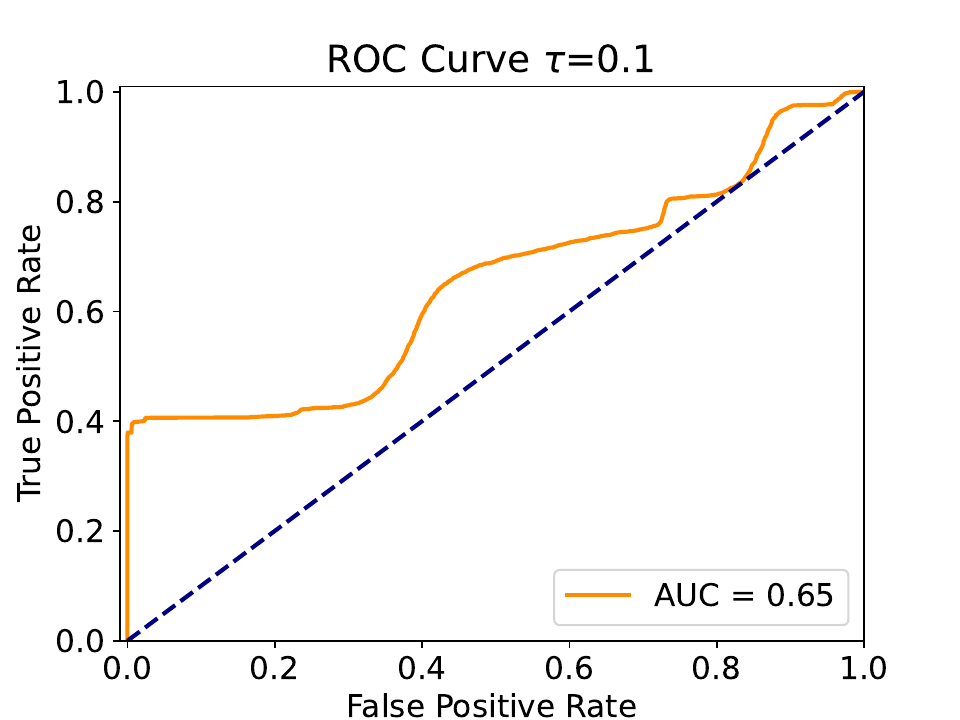}
    \end{subfigure}
    \hfill
    \begin{subfigure}{0.21\linewidth}
    \includegraphics[width=\linewidth]{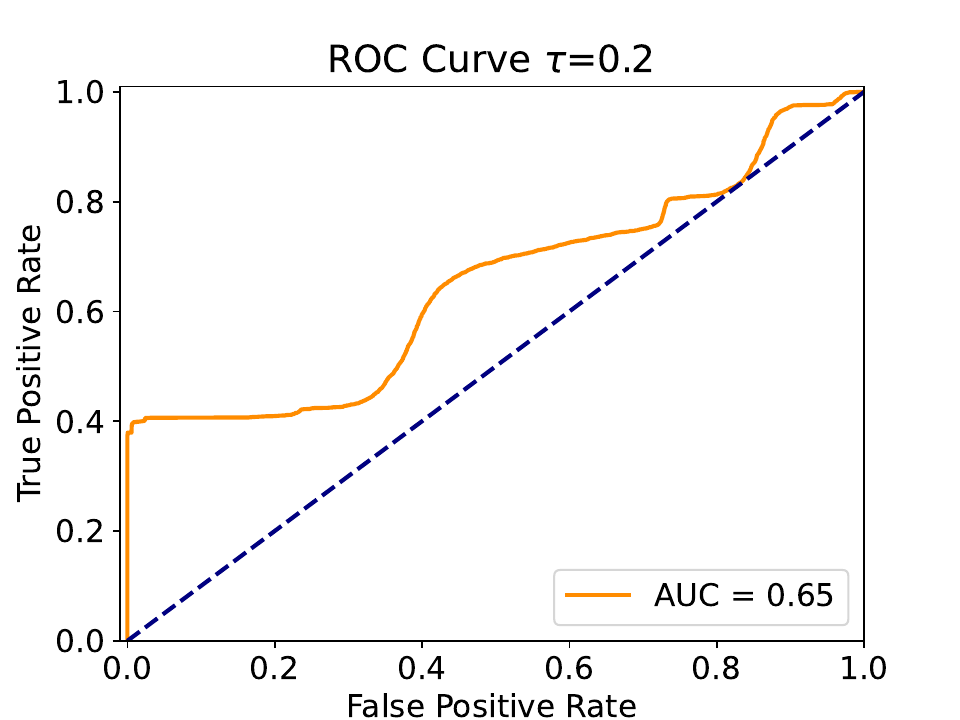}
    \end{subfigure}
    \hfill
    \begin{subfigure}{0.21\linewidth}
    \includegraphics[width=\linewidth]{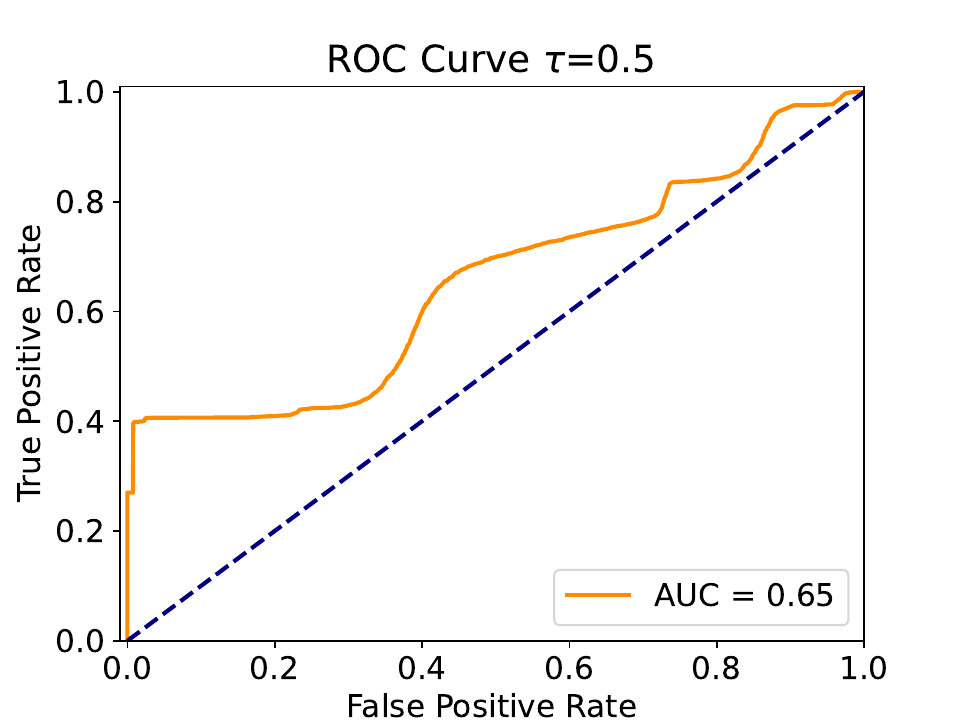}
    \end{subfigure}
    \hfill
    \begin{subfigure}{0.21\linewidth}
    \includegraphics[width=\linewidth]{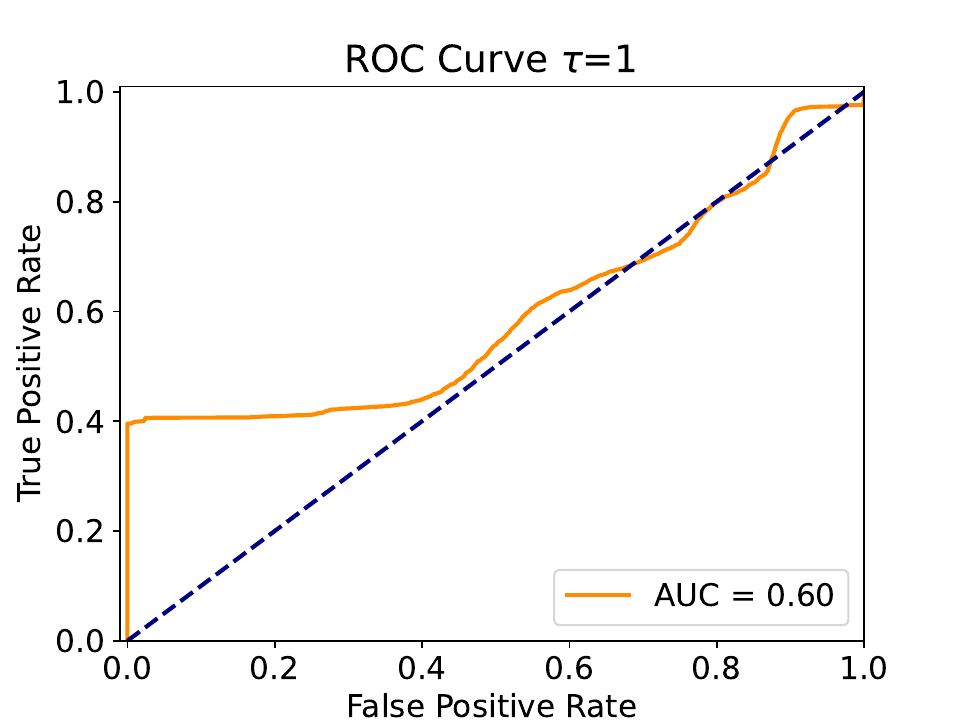}
    \end{subfigure}
    \caption{ROC curves obtained for different values of $\tau$ with method ACGCTNAD without scaler, with training dataset.}
    \label{fig:ROC_Ciber_non_scaler}
\end{figure}

In this first test we will not apply any scaler and we will put together a set of 4990 normal training data with a set of 20000 data to be tested. We can observe in Fig. \ref{fig:ROC_Ciber_non_scaler} the ROC curves obtained for different values of $\tau$, with exceptionally good results around $\tau=0.01$, with an interval of stability between $\tau=0.005$ and $\tau=0.012$, and worse results outside. At the optimal value of $\tau=0.01$ the threshold is at $0.108$, with an AUROC of $0.98$, an accuracy of $97.72\%$ and a confusion matrix of
$\begin{pmatrix}
114788  & 389\\
66 & 4757
\end{pmatrix}$.

\begin{figure}
    \centering
    \begin{subfigure}{0.21\linewidth}
    \includegraphics[width=\linewidth]{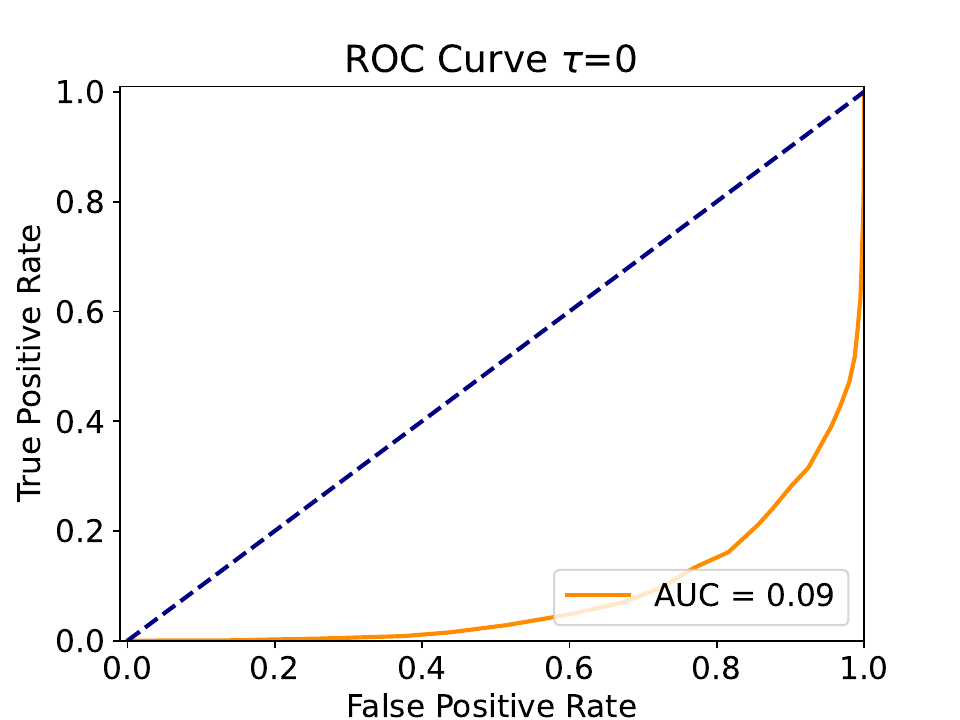}
    \end{subfigure}
    \hfill
    \begin{subfigure}{0.21\linewidth}
    \includegraphics[width=\linewidth]{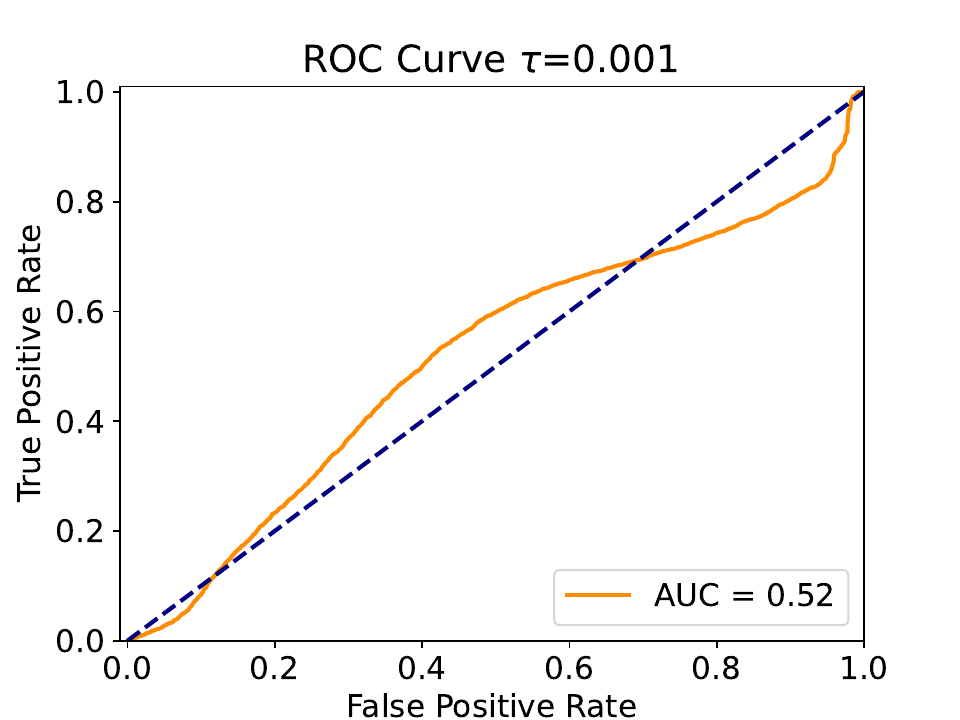}
    \end{subfigure}
    \hfill
    \begin{subfigure}{0.21\linewidth}
    \includegraphics[width=\linewidth]{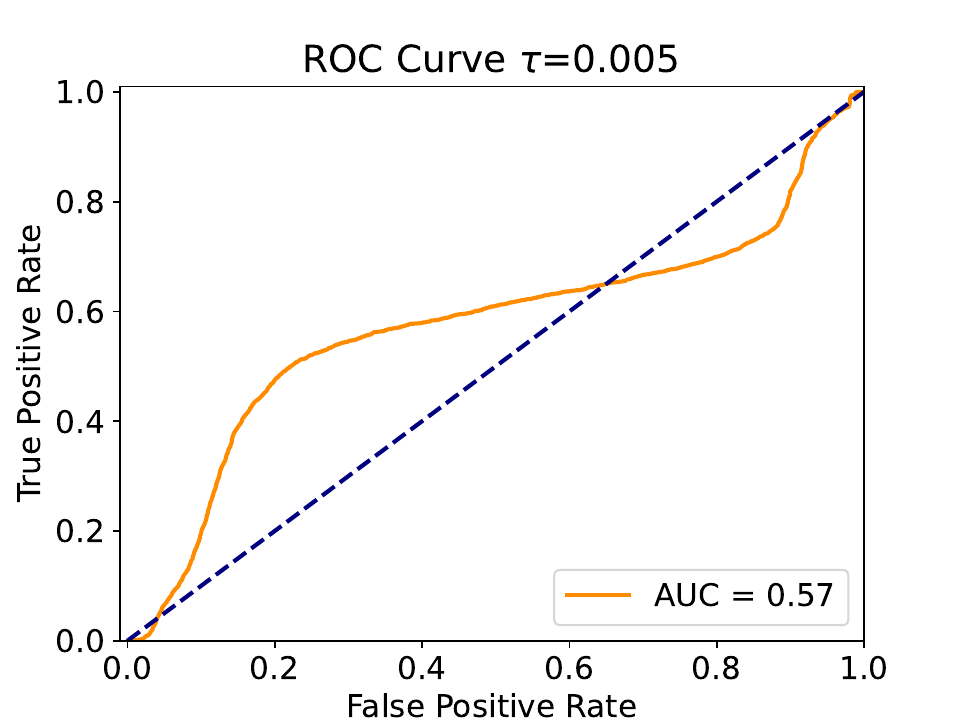}
    \end{subfigure}
    \hfill
    \begin{subfigure}{0.21\linewidth}
    \includegraphics[width=\linewidth]{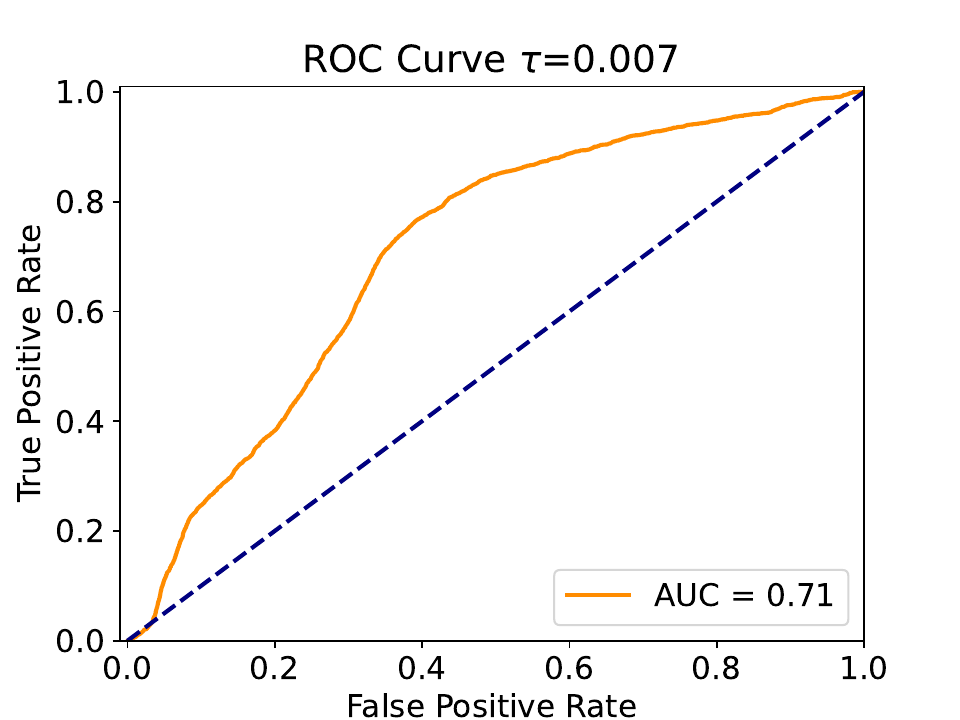}
    \end{subfigure}
    \hfill
    \begin{subfigure}{0.21\linewidth}
    \includegraphics[width=\linewidth]{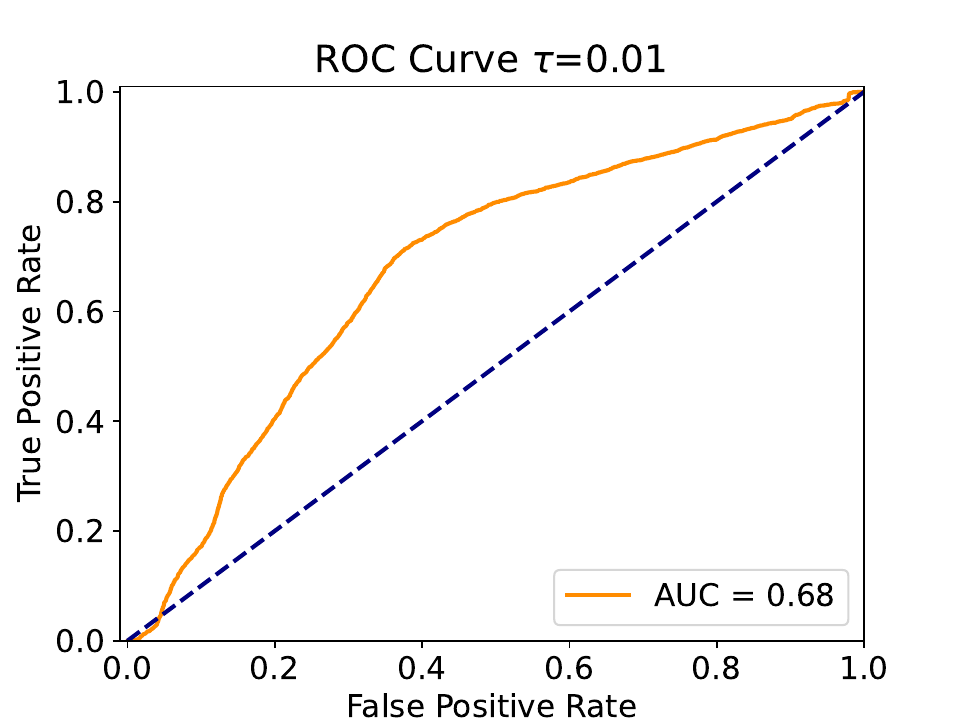}
    \end{subfigure}
    \hfill
    \begin{subfigure}{0.21\linewidth}
    \includegraphics[width=\linewidth]{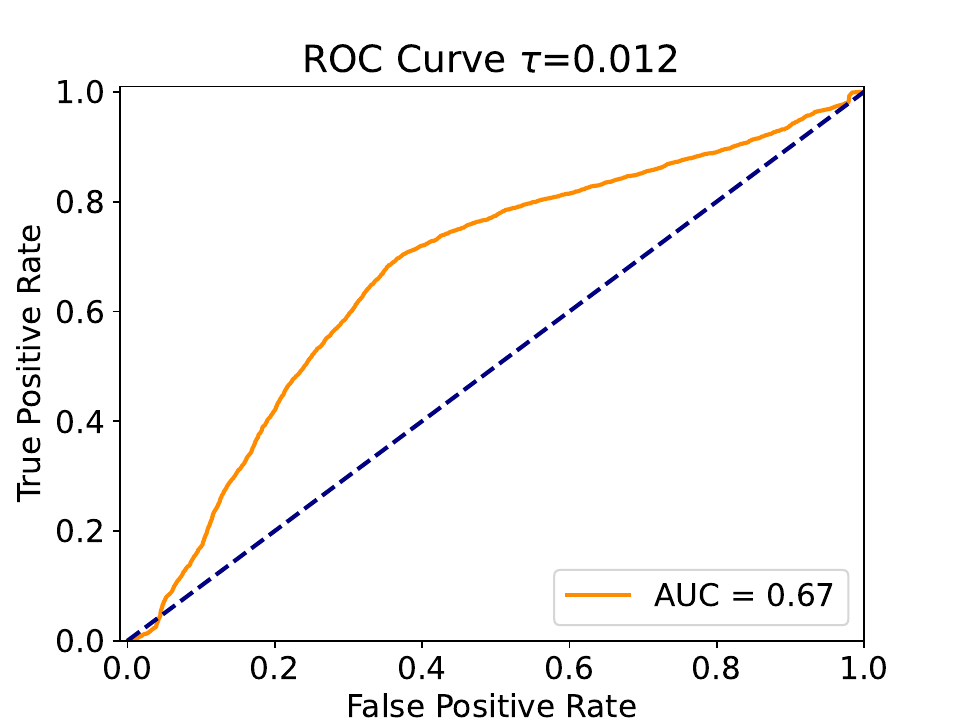}
    \end{subfigure}
    \hfill
    \begin{subfigure}{0.21\linewidth}
    \includegraphics[width=\linewidth]{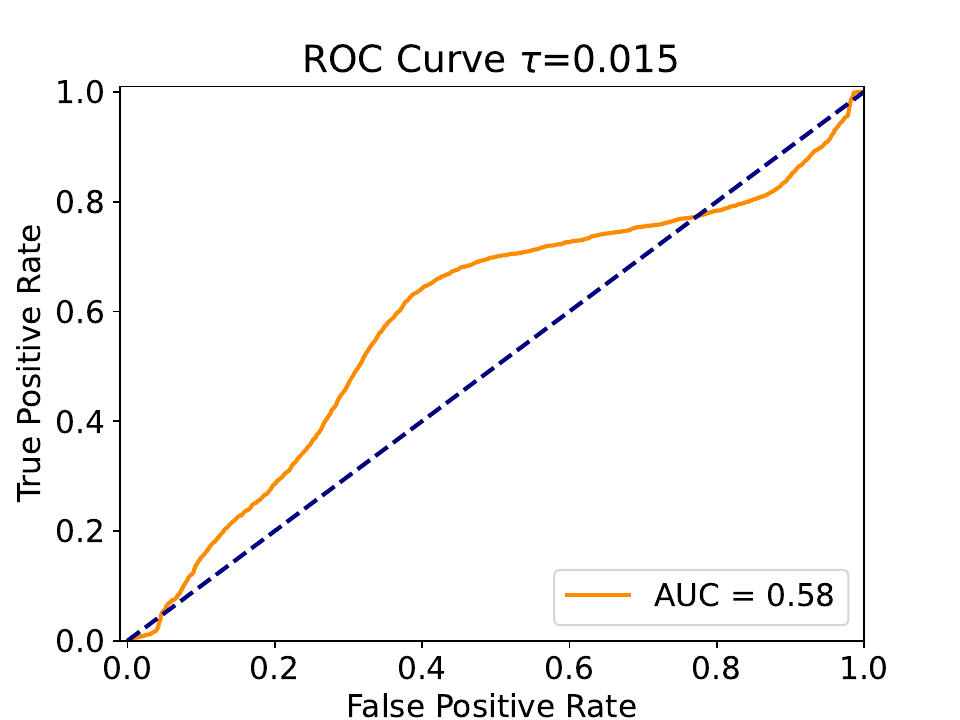}
    \end{subfigure}
    \hfill
    \begin{subfigure}{0.21\linewidth}
    \includegraphics[width=\linewidth]{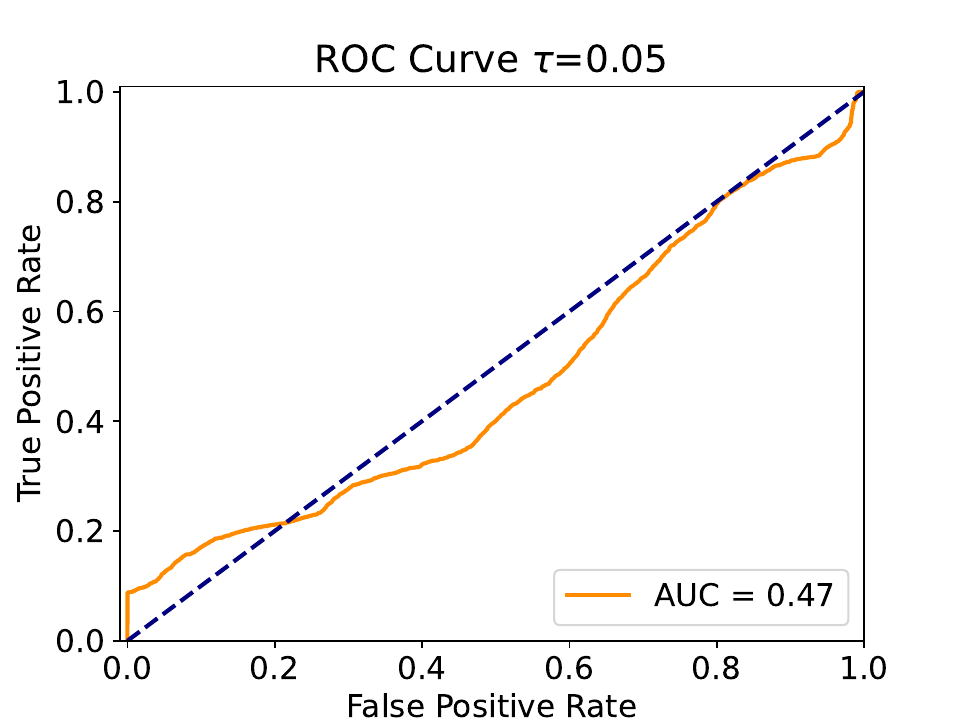}
    \end{subfigure}
    \hfill
    \begin{subfigure}{0.21\linewidth}
    \includegraphics[width=\linewidth]{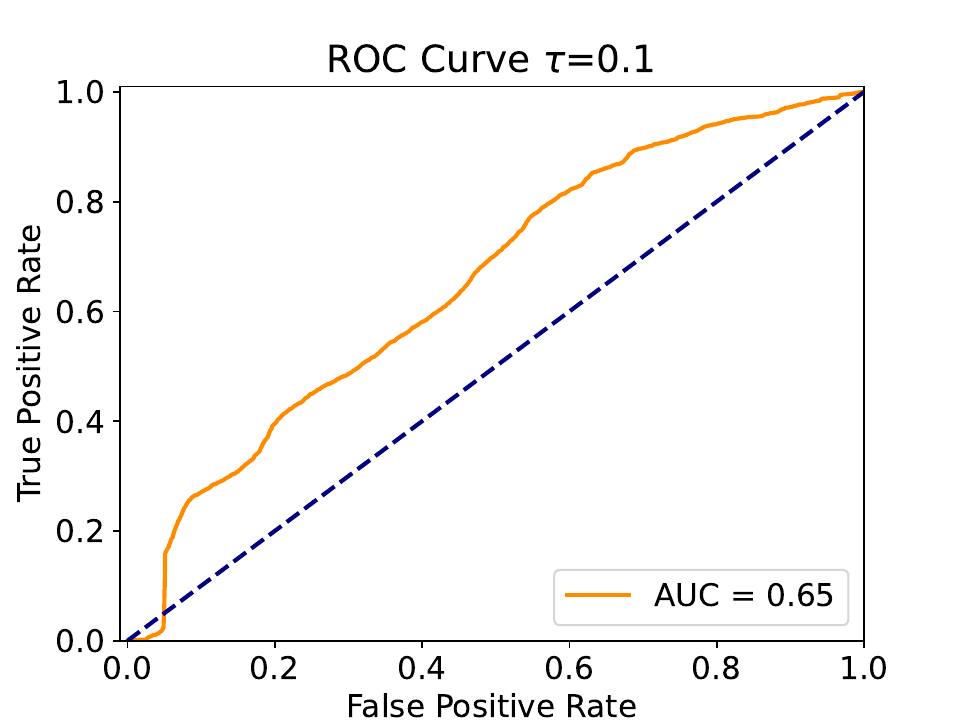}
    \end{subfigure}
    \hfill
    \begin{subfigure}{0.21\linewidth}
    \includegraphics[width=\linewidth]{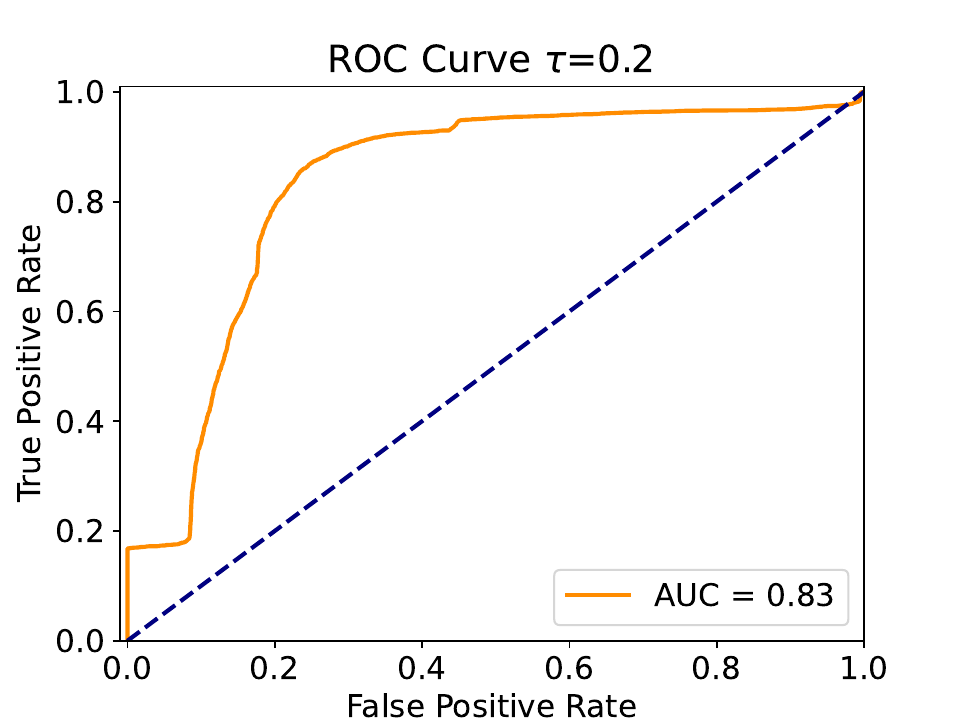}
    \end{subfigure}
    \hfill
    \begin{subfigure}{0.21\linewidth}
    \includegraphics[width=\linewidth]{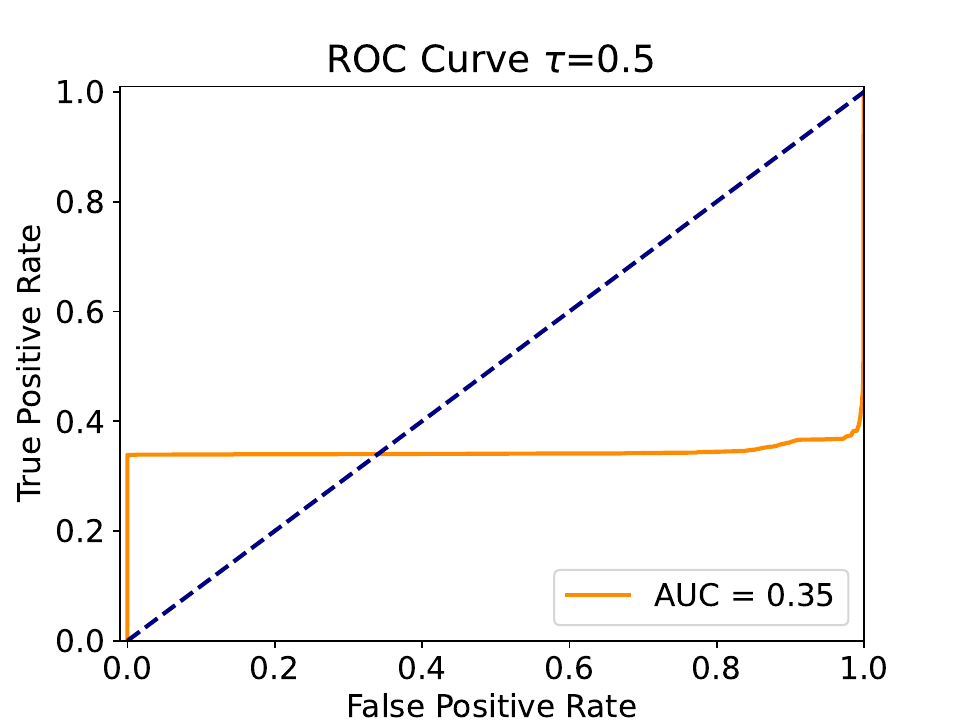}
    \end{subfigure}
    \hfill
    \begin{subfigure}{0.21\linewidth}
    \includegraphics[width=\linewidth]{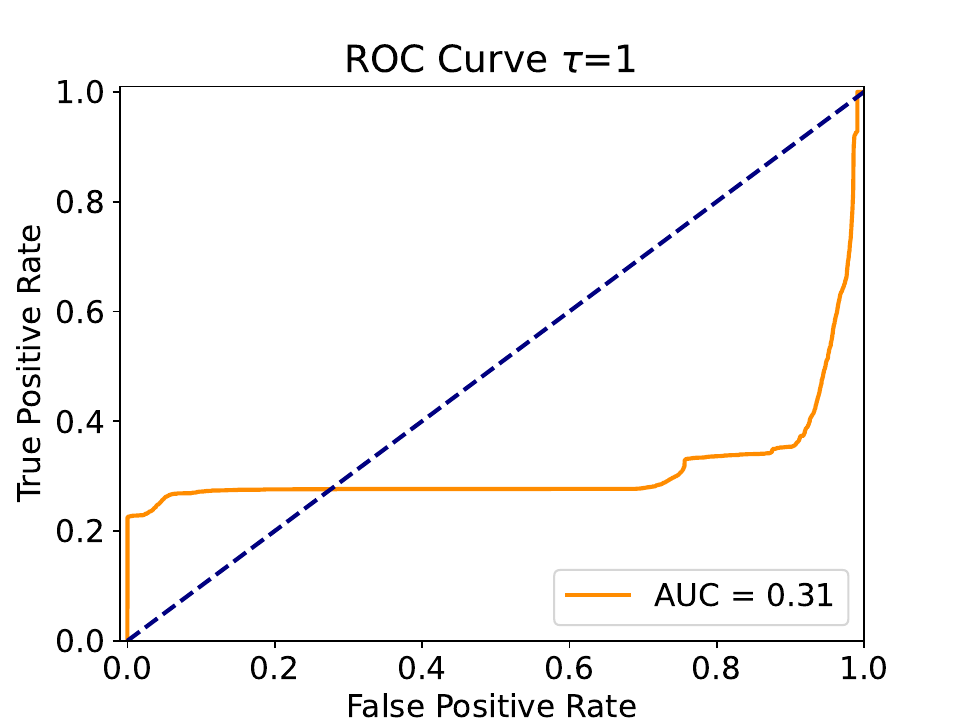}
    \end{subfigure}
    \caption{ROC curves obtained for different values of $\tau$ with method ACGCTNAD with a standard scaler, with training dataset.}
    \label{fig:ROC_Ciber_scaler}
\end{figure}

Now we are going to do the same test by introducing a standard scaler before we input the data in the solver. We can observe in Fig. \ref{fig:ROC_Ciber_scaler} that the result becomes noticeably worse for most cases when we use the scaler, where the best case is that of $\tau=0$, but this is because it is classifying all the data with the same label (understandable knowing that it will return 1 for any data). In all cases, the detector is assigning the same label to almost all the data arbitrarily in the best case, so we see that in this case, applying a scaler ruins the results significantly.

\begin{figure}
    \centering
    \begin{subfigure}{0.45\linewidth}
    \includegraphics[width=\linewidth]{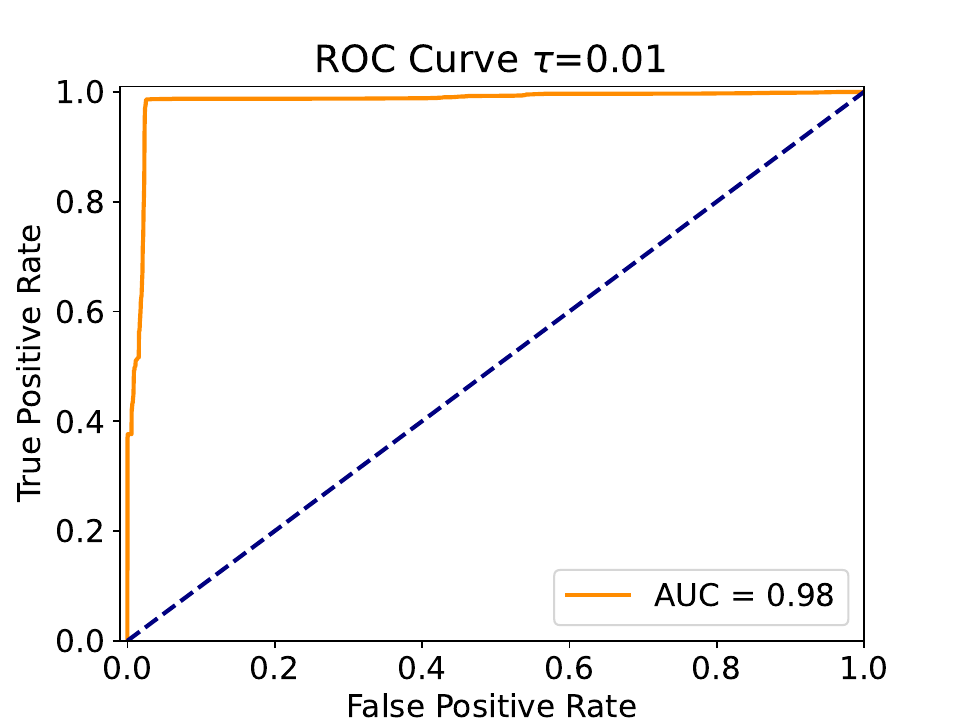}
    \end{subfigure}
    \hfill
    \begin{subfigure}{0.45\linewidth}
    \includegraphics[width=\linewidth]{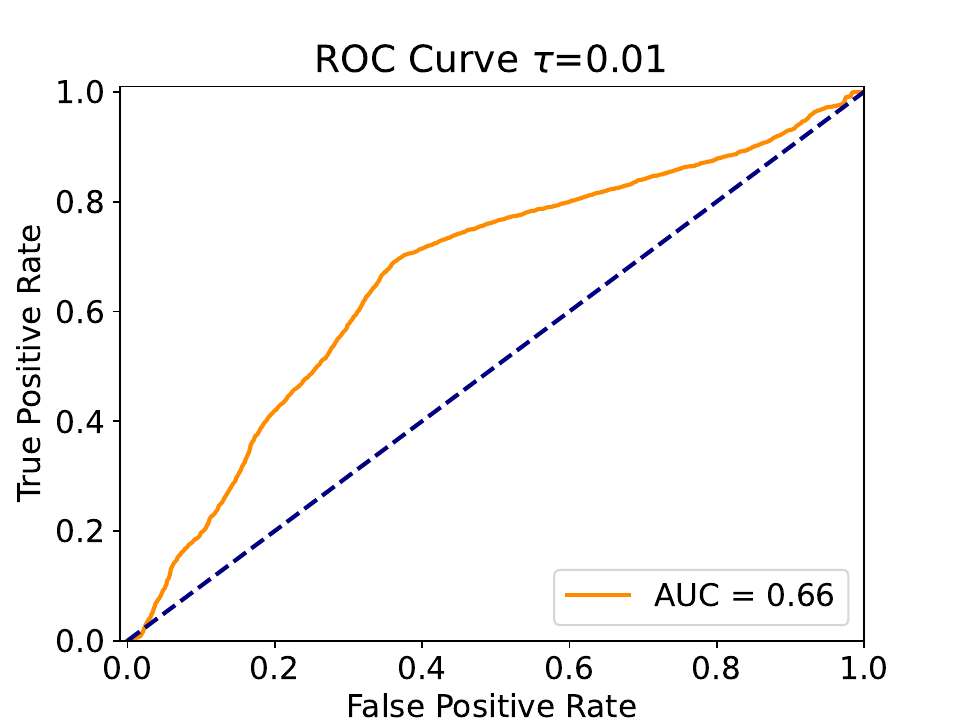}
    \end{subfigure}
    \hfill
    \caption{ROC curves obtained for $\tau=0.01$ with method ACGCTNAD without and with a standard scaler, respectively, without training dataset.}
    \label{fig:ROC_Ciber_scaler_alone}
\end{figure}

Finally, we will test the performance of method ACGCTNAD without including the training dataset, using only data to be tested, both with and without scaler. In Fig. \ref{fig:ROC_Ciber_scaler_alone} we can see the ROC curve results for each one.

In the case without scaler, we can see a very good result, since we have that with a threshold of $0.105$ we have an accuracy of $97.5\%$ and a confusion matrix of
$\begin{pmatrix}
14773   438\\
60  4729
\end{pmatrix}$.
In the case with scaler, the threshold is $0.9999706310280276$ and the accuracy is $64.7\%$, with a confusion matrix
$\begin{pmatrix}
9523 & 5542\\
1500 & 3435
\end{pmatrix}$.

We can easily observe that applying a standard scaler always yields worse results.

\section{Conclusions}
We have created a set of tensor network algorithms based on data compression to detect anomalies in datasets. Among the reported experiments, the local method ACLCTNAD attains the highest peak AUROC values, but at the cost of excessive execution time and with a local derivation that should be interpreted as heuristic. On the other hand, the global method ACGCTNAD is extremely versatile and gives very good results on the tested datasets, which include images and cyber-attack data. We have also observed that for certain situations, applying a pre-scaler to the data can remove all the predictive power of the algorithm. Furthermore, we have seen that the identification power arises from the structure found in the tensor network, and that there is usually a region of maximum identification for a low compression value, which may seem counter-intuitive at first.

A more exhaustive empirical evaluation should report random seeds, the number of runs, and standard deviations, explain how $\tau$ is selected without using test labels, and compare against standard anomaly-detection baselines such as PCA reconstruction error, truncated SVD, Isolation Forest, One-Class SVM, Local Outlier Factor, and autoencoder reconstruction error. In the cybersecurity case, the dataset used here is internal and private, so full public identification is not possible; future reproducibility efforts should therefore describe as much of the acquisition and preprocessing protocol as confidentiality permits, or complement the study with comparable public benchmarks. These additions are left for future work and are not introduced retroactively into the present experimental results.

Future lines of research may be the study of this algorithm with other types of representations, such as PEPS, more suitable for other contexts, such as images, the study of the use of different values of $\tau$ along the truncated SVD, the application to other types of data, such as texts or videos, and the empirical evaluation of the projection-based local variant introduced above.

\bibliographystyle{unsrt}
\bibliography{references}

\end{document}